\documentclass[journal]{IEEEtai}

\usepackage[colorlinks,urlcolor=blue,linkcolor=blue,citecolor=blue]{hyperref}

\usepackage{color,array}

\usepackage{graphicx}
\usepackage{dcolumn}
\newcolumntype{d}[1]{D{.}{.}{#1}}

\newtheorem{theorem}{Theorem}

\usepackage{graphicx}
\usepackage{subfigure}
\usepackage{amssymb}
\usepackage{amsmath}
\usepackage{multicol}
\usepackage{multirow}
\usepackage{bm,xstring}
\usepackage{color}
\usepackage{times}
\usepackage{epsfig}
\let\labelindent\relax
\usepackage[inline]{enumitem}
\usepackage{lipsum}
\usepackage[group-separator={,}]{siunitx}
\usepackage[table,xcdraw]{xcolor}
\usepackage{soul}
\usepackage{makecell}
\usepackage{algorithm}
\usepackage{algorithmic}
\usepackage{wrapfig}
\usepackage{textcomp}
\usepackage{xspace}
\usepackage{verbatim}

\begin{document}

\title{Clustering by the Probability Distributions from Extreme Value Theory}

\author{Sixiao Zheng, Ke Fan, Yanxi Hou,  Jianfeng Feng, and Yanwei Fu, \IEEEmembership{Member, IEEE}

\thanks{Sixiao Zheng is with the Academy for Engineering \& Technology, Fudan University, Shanghai, China (e-mail: sxzheng18@fudan.edu.cn).}

\thanks{Ke Fan,  Yanxi Hou, Jianfeng Feng, and Yanwei Fu are with the School of Data Science, Fudan University, Shanghai, China (e-mail: \{kfan17, yxhou, jffeng, yanweifu\}@fudan.edu.cn). }

\thanks{Jianfeng Feng and Yanwei Fu  are also with Fudan ISTBI—ZJNU Algorithm Centre for Brain-inspired Intelligence, Zhejiang Normal University, Jinhua, China.}

\thanks{Jianfeng Feng is also with the ISTBI, Fudan University, Shanghai, China.}
}

\markboth{Journal of IEEE Transactions on Artificial Intelligence, Vol. 00, No. 0, Month 2020}
{Sixiao Zheng \MakeLowercase{\textit{et al.}}: GPD $k$-means}

\maketitle

\begin{abstract}

Clustering is an essential task to unsupervised learning. It tries to automatically separate instances into ``coherent'' subsets.
As one of the most well-known clustering algorithms,  $k$-means assigns sample points at the boundary to a unique cluster, while it does not utilize the information of sample distribution or density. Comparably, it would potentially be more beneficial to consider the probability of each sample in a possible cluster.
To this end, this paper generalizes $k$-means to model the distribution of clusters. Our novel clustering algorithm thus models the distributions of distances to centroids over a threshold by Generalized Pareto Distribution (GPD) in Extreme Value Theory (EVT). Notably, we propose the concept of centroid margin distance, use GPD to establish a probability model for each cluster, and perform a clustering algorithm based on the covering probability function derived from GPD. Such a GPD $k$-means thus enables the clustering algorithm from the probabilistic perspective.
Correspondingly, we also introduce a naive baseline, dubbed as Generalized Extreme Value (GEV) $k$-means. GEV fits the distribution of the block maxima. In contrast, the GPD fits the distribution of distance to the centroid exceeding a sufficiently large threshold, leading to a more stable performance of GPD $k$-means. Notably, GEV $k$-means can also estimate cluster structure and thus perform reasonably well over classical $k$-means. Thus, extensive experiments on synthetic datasets and real datasets demonstrate that GPD $k$-means outperforms competitors. The github codes are released in \url{https://github.com/sixiaozheng/EVT-K-means}.

 

\end{abstract}

\begin{IEEEImpStatement}
Clustering is an essential task to unsupervised learning. The most well-known clustering algorithm is the k-means. The k-means algorithm assigns each sample to the nearest unique cluster. However, due to the lack of prior information on feature space, if a sample is at the boundary of several clusters, assigning the sample to any one of them may be inappropriate. Instead,
This paper considers the probabilities of the sample in each possible cluster. It is the first time in the literature to generalize k-means by the probabilistic tools in  Extreme Value Theory (EVT). Our novel clustering algorithm thus models the distributions of distances to centroids over a threshold by Generalized Pareto Distribution (GPD).
The GPD k-means we proposed can be widely used in many fields, including customer group analysis, geographic information analysis, network text analysis, e-commerce purchase behavior analysis, and other fields. The GPD k-means will bring some positive effects, such as improving the performance of the clustering task, effectively clustering the streaming data, and improving the performance of the clustering analysis system.
\end{IEEEImpStatement}



\section{Introduction}

\IEEEPARstart{C}{lustering} is an essential task to unsupervised learning \cite{jain2010data,1427769}. It aims at clustering some unlabeled instances with high similarity into one cluster. The most well-known clustering algorithm is the $k$-means~\cite{macqueen1967some}, whose objective is to minimize the sum of the squared distances of the samples to their closest centroid. Notably, the $k$-means algorithm is NP-hard, even when $k$=2~\cite{aloise2009np}. 
The $k$-means algorithm has been extensively studied in the literature, and some heuristics have been proposed to approximate it~\cite{jain2010data, dubes1988algorithms}. The most famous one is Lloyd's algorithm~\cite{lloyd1982least}, due to its simplicity, ease of use, geometric intuition~\cite{bottesch2016speeding}. 
However, Lloyd's algorithm uses a group of randomly initialized centroids, with no guarantee that the objective function reaches the global minimum. The $k$-means++ algorithm~\cite{arthur2007k} was proposed to solve this problem by using an adaptive sampling scheme called $D^2$-sampling to find a good initialization for the centroids.

The $k$-means algorithm assigns each sample to the nearest unique cluster.
However, due to the lack of prior information on feature space, if a sample is at the boundary of several clusters, assigning the sample to any one of them may be inappropriate. Instead, it might be more interesting to consider the probabilities of the sample in each possible cluster.

The fuzzy $c$-means algorithm~\cite{dunn1973fuzzy,bezdek2013pattern} is amenable to alleviating  this problem, as it allows each sample to have memberships in all clusters rather than attaching to a particular cluster. Nevertheless, the fuzzy c-means algorithm only directly relaxes the `cluster loyalty' of each sample to a value between zero and one, rather than purely understanding the clustering from a probabilistic perspective. To this end, this paper studies the probability-based $k$-means clustering algorithm.

In this paper, we first introduce a Generalized Extreme Value (GEV) $k$-means clustering algorithm based on Extreme Value Theory (EVT)~\cite{coles2001introduction}, and then in order to make full use of the extreme value information in the data, we propose another novel $k$-means algorithm based on Generalized Pareto Distribution (GPD) by establishing a probability model for each cluster based on EVT.  
As a branch of statistics, the EVT is applied to model the stochastic behavior of the extreme samples found in the distribution tail~\cite{coles2001introduction}. 
To model a cluster, we present the novel concept of~\emph{centroid margin distance},  defined as the minimum pairwise distance between a centroid and the samples from other clusters. 
To facilitate modeling, we do not directly fit the distribution of the centroid margin distance but the distribution of the negative centroid margin distance. The distribution of the negative centroid margin distance shall be approximated by a GEV~\cite{jenkinson1955frequency} distribution or GPD~\cite{pickands1975statistical}. From the GEV or GPD, we can then derive a \emph{covering probability} function, which indicates the probability that a sample is covered by a cluster. 
The larger the covering probability, the higher probability the sample belongs to the cluster. 

The objective of GEV $k$-means and GPD $k$-means is to minimize the sum of negative covering probability. 
For the GEV $k$-means, we first assign group labels to each sample (to distinguish from cluster labels) and select the maximum negative pairwise distance within each block. We then apply the Block Maxima Method (BMM) to fit a GEV distribution for each cluster by these distances. Then each sample is assigned to a cluster by the maximum covering probability. Finally, we update centroids to the mean of all samples in the cluster. These three steps are iteratively computed until the centroids no longer change.
However, BMM only uses a minimal amount of negative pairwise distance, resulting in a significant waste of data. 
Unlike the GEV $k$-means, the GPD $k$-means makes full use of the extreme value information in the data by using the Peaks-Over-Threshold (POT) method \cite{leadbetter1991basis} to model the excess of negative pairwise distance exceeding a chosen threshold and fit a GPD for each cluster, and the other steps are the same as GEV $k$-means.  

This paper makes the following contributions:
(1) We generalize the $k$-means to model the cluster by GPD $k$-means, which is proposed to enable the clustering from the probabilistic perspective. 
(2) We propose the concept of centroid margin distance, and use the GPD to fit the negative centroid margin distance, thereby deriving the covering probability function for assigning samples.
(3) We also introduce a na\"ive baseline for GPD $k$-means, namely, GEV $k$-means.
(4) Extensive experimental results show that the GPD $k$-means outperforms the competitors consistently across all experimental datasets. Note that our GPD $k$-means is very robust to uninformative features. With the increase of the number of  uninformative features, our GPD $k$-means does not decline, but has a slight upward trend.

\section{Related Works}
The $k$-means has been extensively studied in the literature in many aspects \cite{jain2010data, 1427769}.
The basic $k$-means have been expanded into many successful algorithms by different methods. We can only highlight some of these works here.
The $k$-means++ \cite{arthur2007k} is the most popular initialization scheme to provide a good initialization for centroids.
Fuzzy $c$-means proposed by \cite{dunn1973fuzzy} and later modified by \cite{bezdek2013pattern}, is an extension of $k$-means where each sample has memberships in all clusters. 
In \cite{karypis2000comparison}, the authors proposed a hierarchically divisive version of $k$-means, called bisecting $k$-means, that recursively partitions the data into two clusters at each step.
In $k$-medoid \cite{kaufman2009finding}, clusters are represented using the median of the data instead of the mean. Kernel $k$-means \cite{scholkopf1998nonlinear} was proposed to detect arbitrarily shaped clusters, with an appropriate choice of the kernel similarity function.
DBSCAN \cite{ester1996density} searches connected dense regions by computing the density of samples to perform cluster assignments.
Gaussian Mixture Model (GMM) \cite{bishop2006pattern} 
assumes that the sample is generated from a mixture of Gaussian distribution and estimates the parameters of GMM for clustering. In contrast, the GEV $k$-means and GPD $k$-means only fit the negative centroid margin distance, and the extreme value distribution is more complicated and significantly different from the Gaussian distribution.

EVT has been widely applied in natural phenomena, finance, traffic prediction, and other fields. In recent years, there have been an increasing number of applications in machine learning related to EVT \cite{Rudd_2018_TPAMI,Scheirer_2017_MC,Scheirer_2014_TPAMIb,Jain_2014_ECCV}. 
However, less attention is paid to exploiting EVT to improve $k$-means. 
In \cite{li2012feature}, the authors proposed using GEV distribution for feature learning based on $k$-means. However, our method is significantly different from this method. First, they compute the squared distance from a point to the nearest centroid and form a GEV regarding each point, while we compute centroid margin distance and use GEV and GPD to establish a probability model for each cluster. Second, their algorithm adds the likelihood function as a penalty term into the objective function of $k$-means; in contrast, the objective of GEV $k$-means and GPD $k$-means is to minimize the sum of the negative covering probability, which indicates the probability that a sample is covered by a cluster. Finally, we also propose a GPD-based $k$-means algorithm, which was not mentioned in \cite{li2012feature}.

\section{Preliminaries}

\subsection{Extreme Value Theory} 
We introduce GEV and GPD derived from two theorems.
\begin{theorem}[Fisher-Tippett Theorem]
\textnormal{\cite{fisher1928limiting}}

\noindent Let $X_1, X_2,\ldots, X_n$ be a sequence of independent and identically distributed (i.i.d.) random variables with distribution $F$.
Let $M_n=\max_{1\le i\le n} X_i$ as the sample maximum. If there exist sequences of constants $a_n>0$ and $b_n$ such that
\begin{equation}
   \lim _{n \to \infty} P\left(\frac{M_n-b_n}{a_n} \leqslant x\right) \to H(x), \quad x \in \mathbb{R},
   \label{equ:Fisher_Tippett}
\end{equation}
then if $H(x)$ is a non-degenerate distribution function, $H$ must belong to the family of Generalized Extreme Value (GEV) distributions with 
\begin{equation}
   H(x)=\exp\left\{-\left(1+\xi\frac{x-\mu}{\sigma}\right)^{-1/\xi}\right\},
   \label{equ:GEV}
\end{equation}
where $1+\xi\frac{x-\mu}{\sigma} > 0$. 
$\mu,\xi \in \mathbb{R}$ and $\sigma>0$ are the location, shape and scale parameters, respectively.
\label{theorem:FT}
\end{theorem}
\begin{theorem}[Pickands-Balkema-de Haan  Theorem]\textnormal{\cite{balkema1974residual}}
Let $X_1, X_2,\ldots, X_n$ be a sequence of independent and identically distributed (i.i.d.) random variables, and the right end-point supported by the distribution function $F$ is $x^{*}$. Assuming that there is a sufficiently large threshold $u (u<x^*)$, $X_i-u$ is the excess, then the cumulative distribution function of the excess $F_u$ is
\begin{equation}\label{equ:threshold}
    F_u(x)=P(X-u\leqslant x|X>u)=\frac{F(x+u)-F(u)}{1-F(u)},
\end{equation}
where $x \geqslant 0$. $F_u$ can be approximated using Generalized Pareto Distribution (GPD) with
\begin{equation}\label{equ:GPD}
    G(x)=1-\left(1+\xi\frac{x-\mu}{\sigma}\right)^{-1/\xi},
\end{equation}
where $x \geqslant \mu, 1+\xi\frac{x-\mu}{\sigma}>0$. 
$\mu,\xi \in \mathbb{R}$ and $\sigma>0$ are the location, shape and scale parameters, respectively.
\label{theorem:PBH}
\end{theorem}

\subsection{$k$-means Clustering}
Denote $\mathcal{X}=\{\bm{x}_1,\bm{x}_2,\dots,\bm{x}_n\} \subseteq \mathbb{R}^d$ as the dataset and $\mathcal{C}=\{C_1,C_2,\ldots,C_k\}$ as a partition of $\mathcal{X}$ satisfying $C_i\cap C_j=\varnothing, i\neq j$. Let $\Theta=\{\bm{\theta}_1,\bm{\theta}_2,\ldots,\bm{\theta}_k\}$ with
$\bm{\theta}_j \in \mathbb{R}^d$ being the centroid of cluster $C_j, j=1,2,\ldots,k$, that is, $\bm{\theta}_j = \frac{1}{|C_j|} \sum_{\bm{x} \in C_j}\bm{x}$.
$k$-means can be cast as minimization of the objective function
\begin{equation}\label{equ:$k$-means_objective}
    J(\mathcal{C};\Theta) = \sum_{j=1}^{k} \sum_{\bm{x} \in C_j}\|\bm{x}-\bm{\theta}_j\|^2_2.
\end{equation}
Lloyd's algorithm \cite{lloyd1982least} uses a greedy strategy to approximate the $J(\mathcal{C};\Theta)$ by iteratively optimizing between assigning samples  to their closest centroid and updating each centroid by averaging over its assigned samples. 
These two steps loop iteratively until the centroids no longer change.

\section{$k$-means via Extreme Value Theory \label{sec:3}}
\subsection{Centroid Margin Distance Distribution\label{sec:4.1}}
\begin{figure}[t]
\centering
\includegraphics[width=1.0\columnwidth]{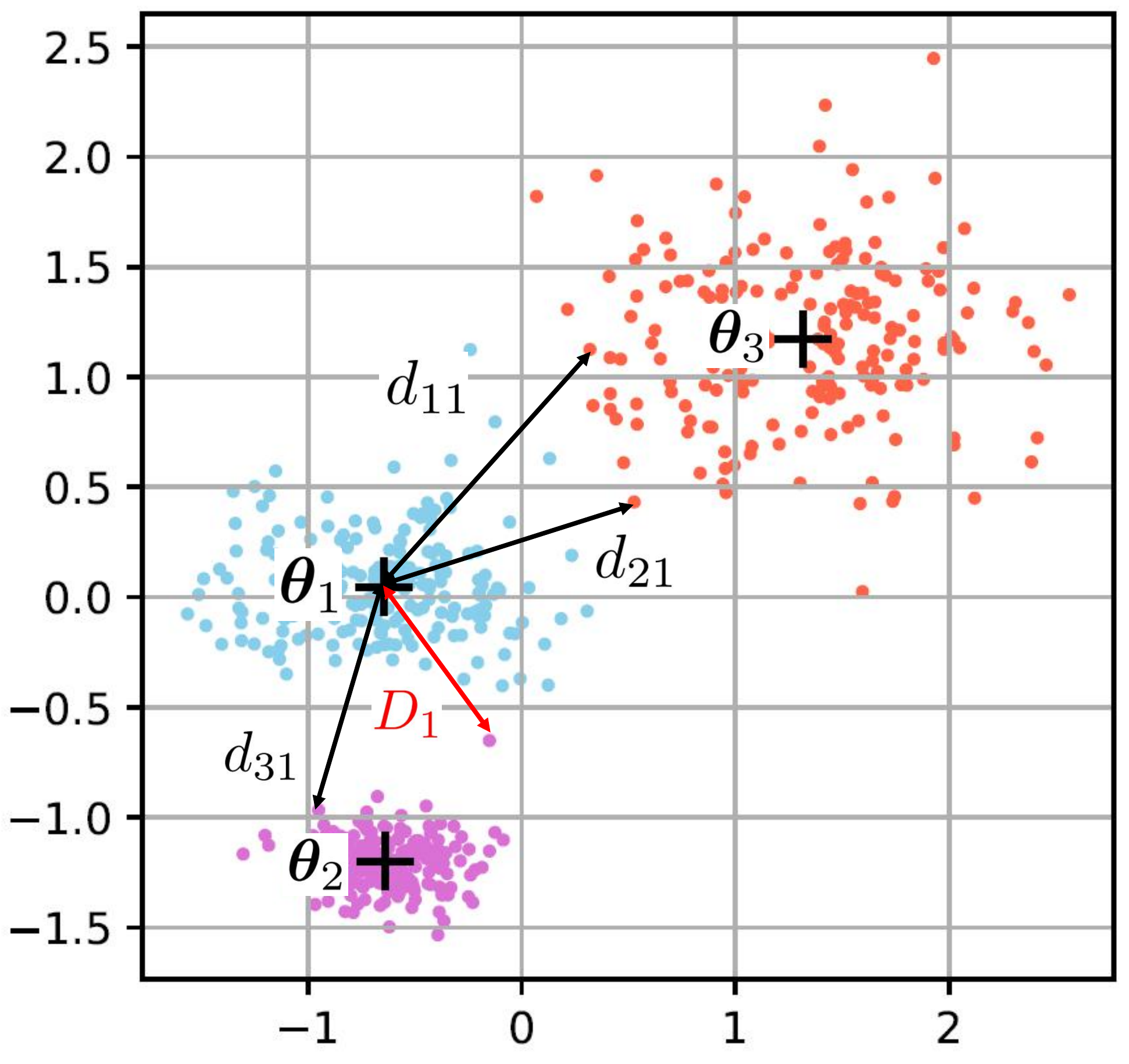}
\caption{The centroid margin distance of centroid $\bm{\theta}_1$}
\label{fig:em_dist}
\end{figure}

We propose the concept of \emph{centroid margin distance} to help model the cluster from a probabilistic perspective. Particularly, it is necessary to know the sample distribution of clusters and the relationship between clusters. 
Therefore, the centroid margin distance is defined as the minimum pairwise distance between a centroid and the samples of other clusters, as shown in $D_1$ in Fig. \ref{fig:em_dist}. Formally, for a cluster $C_j$, given the centroid $\bm{\theta}_j$, the centroid margin distance is defined as $D_j=\min_{i:\bm{x}_i \notin C_j} d_{ij} = \min_{i:\bm{x}_i \notin C_j} \|\bm{x}_i-\bm{\theta}_j\|_2$.
The centroid margin distance is somewhat similar to the margin distance defined in the previous works \cite{Rudd_2018_TPAMI}, which is the half distance from a positive sample to the 
nearest negative samples. 
For a centroid $\bm{\theta}_j$, we can calculate a set of the pairwise distance between $\bm{\theta}_j$ and the samples of other clusters with the minimal values $D_j$.
However, since Theorems \ref{theorem:FT} and \ref{theorem:PBH} are used to fit the distribution of the sample maxima, in order to fit the distribution of the sample minima, we can take the maximal negative centroid margin distance 
\begin{equation}\label{equ:ncmd}
    D'_j = \max_{i:\bm{x}_i \notin C_j} \{-d_{ij}\} = \max_{i:\bm{x}_i \notin C_j} -\|\bm{x}_i-\bm{\theta}_j\|_2,
\end{equation}
as $\min_{i:\bm{x}_i\notin C_j} d_{ij}=-\max_{i:\bm{x}_i\notin C_j}\{ -d_{ij}\}$.
The distribution of $D'_j$ can be approximated by an extreme value distribution (GEV or GPD) with suitable methods. From the extreme value distribution, we can then derive a \emph{covering probability} function, which indicates the probability that a sample is covered by a cluster. 

When applying the GEV, the probability that sample $\bm{x}$ is covered by the cluster $C_j$ is
\begin{equation}
\begin{aligned}
   P(\bm{x},\bm{\theta}_j)&=H(-\|\bm{x}-\bm{\theta}_j\|_2) \\
   &=\exp\left\{-\left(1+\xi\frac{-\|\bm{x}-\bm{\theta}_j\|_2-\mu}{\sigma}\right)^{-1/\xi}\right\}.
\end{aligned}
   \label{equ:P_GEV}
\end{equation}

Similarly, when applying the GPD, the probability that sample $\bm{x}$ is covered by the cluster $C_j$ is
\begin{equation}
\begin{aligned}
    P(\bm{x},\bm{\theta}_j)&=G(-\|\bm{x}-\bm{\theta}_j\|_2)\\
    &=1-\left(1+\xi\frac{-\|\bm{x}-\bm{\theta}_j\|_2-\mu}{\sigma}\right)^{-1/\xi}.
\end{aligned}
    \label{equ:P_GPD}
\end{equation}
The location $\mu$, scale $\sigma$, and shape $\xi$ parameters of GEV and GPD are obtained by fitting the maximal negative centroid margin distance through the widely used Maximum Likelihood Estimation (MLE). 

\subsection{Naive Approach: GEV $k$-means Algorithm \label{sec:3.3}}
As described in Sec. \ref{sec:4.1}, the larger the covering probability, the higher the probability sample $\bm{x}$ is covered by the cluster. Therefore, We propose GEV $k$-means based on the covering probability, which is summarized in Alg. \ref{alg:GEV_kmeans}. 
The GEV $k$-means minimizes the objective function of the sum of negative covering probability. 
\begin{equation}\label{equ:GEV_obj}
    J'(\mathcal C;\Theta)=\sum_{j=1}^{k} \sum_{\bm{x} \in C_j}\left(-P(\bm{x},\bm{\theta}_j)\right).
\end{equation}
Note that the covering probability $P$ is associated with some unknown parameters as in Eq. \eqref{equ:P_GEV} and \eqref{equ:P_GPD}, so we introduce an MLE step in our Algorithms 1 and 2 to update the parameters in each iteration. 
Since the MLE method naturally possesses statistical consistency, it effectively learns the covering probabilities.
More details are discussed in Sec. \ref{sec:4.4} and Sec. \ref{sec:4.5}. In addition, each centroid has only one closest sample of other clusters, that is, $D'_j$ has only one observed sample, so we should consider sampling multiple $D'_j$ observed samples to fit the GEV. 

\begin{figure*}[t]
\centering
\includegraphics[width=1.0\linewidth]{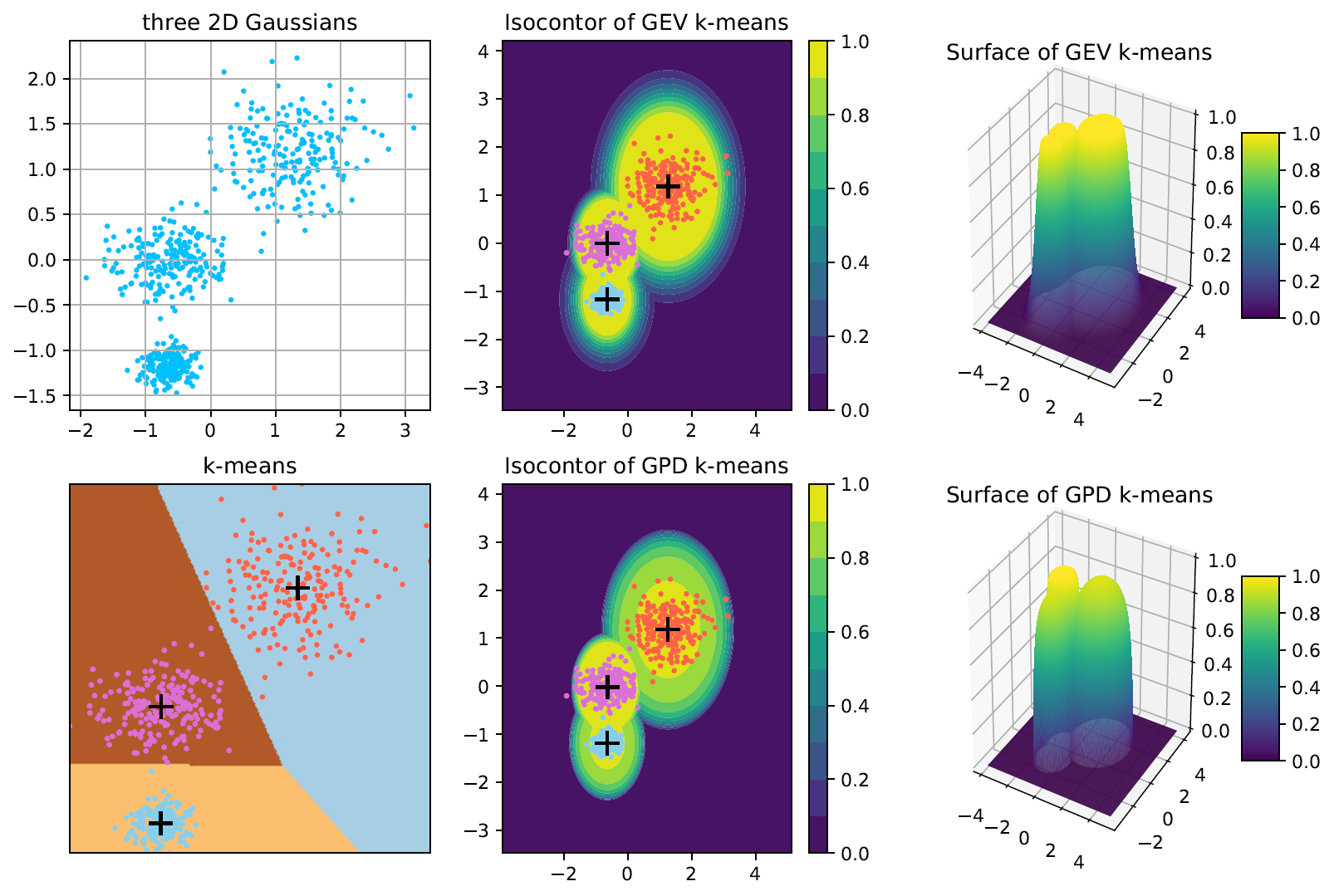}
\vspace{-0.15in}
\caption{Clustering results on three 2D Gaussian data. \textbf{Three 2D Gaussian}: raw data. \textbf{$k$-means}: $k$-means clustering result and decision boundaries. \textbf{Isocontor of GEV $k$-means, Surface of GEV $k$-means}: clustering result with contour plots and surface plots of GEV $k$-means covering probability. \textbf{Isocontor of GPD $k$-means, Surface of GPD $k$-means}: clustering result with contour plots and surface plots of GPD $k$-means covering probability.}
\label{fig:evtkmeans_kmeans}
\end{figure*}

Theorem \ref{theorem:FT} motivates the BMM \cite{gumbel2012statistics}, which uses GEV to provide an ideal model for the maxima of blocks with equal size $s$.
The BMM can be applied to fit the GEV distribution for each cluster to compute the covering probabilities. We first compute the Euclidean distance $d_{ij}^o$ between $\bm{\theta}_j$ and sample $\bm{x}_i \in \mathcal{X}$, \emph{i.e.}, $d_{ij}^{o}=\|\bm{x}_i-\bm{\theta}_j\|_2$. 
Then a group label (to distinguish from the cluster label) is assigned to each sample according to the closest centroid. And we have $k$ sample groups $\{G_1, G_2, \ldots, G_k\}$. For the centroid $\bm{\theta}_j$, its pairwise distance between it and samples of other clusters is $d_{ij} = \|\bm{x}_i-\bm{\theta}_j\|_2, \bm{x}_i \notin G_j$. 
Then we use BMM to divide its negative pairwise distance $-d_{ij}$ (Sec. \ref{sec:4.1}) equally into $m$ blocks of size $s$ (possibly the last block with no sufficient observations), and then the maximum value of each block is taken to obtain the block maximum sequence $\bm{M}^j$.
\begin{equation}
   \bm{M}^j=\{M^j_1, M^j_2, \dots, M^j_{m}\}.
   \label{equ:BMM}
\end{equation}

$\bm{M}^j$ can be viewed approximately as multiple observed samples of $D'_j$. We can use $\bm{M}^j$ to estimate the parameters of GEV distributions for centroid $\theta_j$ using MLE. So each centroid has its own independent GEV distribution. 
In assigning cluster labels, each sample is assigned a cluster label based on the maximum covering probability, \emph{i.e.}, $\lambda_i = \mathop{\arg\max}_{j \in \{1,2,\dots, k\}} P(\bm{x}_i,\bm{\theta}_j)$. 
In updating centroid, each centroid is updated to the mean of all samples in the cluster, \emph{i.e.}, $\bm{\theta}_j = \frac{1}{|C_j|} \sum_{\bm{x} \in C_j}\bm{x}$. These three steps loop iteratively until the centroids no longer change.

\subsection{GPD $k$-means Algorithm \label{sec:3.3}}
According to Theorem \ref{theorem:FT}, when the block size $s$ is large enough, $\bm{M}^j$ can be approximately regarded as an independent and identically distributed observation from the GEV distribution. However, when block size $s$ is large enough, BMM only uses a very small amount of negative pairwise distance, resulting in a large waste of data. Furthermore, there may be cases where the second largest value of one block is larger than the maximum value of the other block, which cannot be utilized. 
In order to make full use of the extreme value information in the data, the POT approach focuses on the excess over a large threshold to fit the GPD and asymptotically
characterize the tail features of the distribution, instead of considering the maxima of blocks like BMM. 

Theorem \ref{theorem:PBH} gives rise to the POT approach \cite{pickands1975statistical} that focuses on the excess over the threshold $u$ to fit the GPD. Therefore, we propose GPD $k$-means based on POT approach. Its objective function is the same as that of GEV $k$-means, \emph{i.e.}, $J'(\mathcal C;\Theta)=\sum_{j=1}^{k} \sum_{\bm{x} \in C_j} -P(\bm{x},\bm{\theta}_j)$. The algorithm process of GPD $k$-means is similar to that of GEV $k$-means, as shown in Alg. \ref{alg:GPD_kmeans}. 
We first compute Euclidean distance $d_{ij}^{o}=\|\bm{x}_i-\bm{\theta}_j\|_2$, and obtain $k$ sample groups $\{G_1, G_2, \ldots, G_k\}$. 
Then we use the POT method to model the excess of negative pairwise distance $-d_{ij} $ exceeding threshold $u_j$ for centroid $\bm{\theta}_j$ and fit the GPD. 
The excess is defined as 
\begin{equation}
   \bm{y}^{j}=-d_{ij}-u_j,~ -d_{ij}>u_j, ~\bm{y}^j=\{y^j_1,y^j_2, \dots,y^j_{o_j}\},
   \label{equ:y}
\end{equation}
where $o_j$ is the total number of observations greater than the threshold $u_j$.
Here $u_j$  is the threshold which we manually designed to filter the value below it.
$\bm{y}^{j}$ can be viewed approximately as multiple observed samples of $D'_j$. Then the estimated parameters of GPD for centroid $\theta_j$ are obtained on $\bm{y}^j$ using MLE. Similar to GEV $k$-means, each centroid has its own independent GPD distribution. 
The two steps of assigning cluster label and updating centroid are the same as those of GEV $k$-means, and not be repeated here.

\noindent \textbf{Remark.}
We give some theoretical explanation about the relation and difference between GEV $k$-means and GPD $k$-means here.  
GEV $k$-means and GPD $k$-means both model negative centroid margin distance by the extreme value theory, and thus can well cluster the data in principle. On the other hand, these two models differ in the specific distribution forms. In this paper, we take generalized extreme value distribution for GEV k-means and generalized Pareto distribution for  GPD k-means. GEV applies the block maxima method, while GPD employs the peak-over-threshold method. Generally, estimation of GPD is more effective than GEV.

As shown in Fig. \ref{fig:evtkmeans_kmeans}, GEV $k$-means and GPD $k$-means establish a covering probability model for each centroid. The closer to the centroid, the higher the covering probability. 
It can be clearly seen that the decision boundary between GEV $k$-means or GPD $k$-means and $k$-means is very different. It can be interpreted that the decision boundaries of GEV $k$-means and GPD $k$-means are contour lines with a covering probability of zero. In contrast, the decision boundary of $k$-means is a straight line.

\begin{algorithm}[t]
\caption{GEV $k$-means}              
\label{alg:GEV_kmeans}
{\small
\begin{algorithmic}
  \STATE {\bfseries Input:} $\mathcal{X} \subseteq \mathbb{R}^{d}$, number of cluster $k$, block size $s$.
  \STATE {\bfseries Output:} centroid $\Theta$
  \STATE {\bfseries Initialization:} random centroid $\Theta$
  \REPEAT
    \STATE Compute the Euclidean distance $d_{ij}^o$ and obtain $\{\bm{G}_1, \bm{G}_2, \ldots, \bm{G}_k\}$;
    \FOR{$j=1,2,\dots,k$}
    \STATE Obtain $\bm{M}^j$ by Eq. \eqref{equ:BMM};
    \STATE Estimate the parameters of GEV on $\bm{M}^j$ by MLE;
    \ENDFOR
    
    \FOR{$i=1,2,\dots,n$}          
    \STATE $\lambda_i = \arg\max_{j \in \{1,2,\dots, k\}}P_{ij}$;
    \ENDFOR
    
    \FOR{$j=1,2,\dots,k$}          
    \STATE $\bm{\theta}_j = \frac{1}{|C_j|} \sum_{\bm{x} \in C_j}\bm{x}$;
    \ENDFOR
    
  \UNTIL{centroids no longer change}
        \STATE {\bfseries return} centroid $\Theta$\;  
\end{algorithmic} }
\end{algorithm}

\subsection{Estimate the Parameters of GPD by MLE} \label{sec:4.4}
MLE is a classic method to estimate probability distribution parameters based on samples. Consider a dataset $\mathcal{X}=\{\bm{x}_1,\bm{x}_2,\dots,\bm{x}_n\}$ containing $n$ samples, it is drawn independently from the density function $p(x;\theta)$ parametrized by $\theta$. The maximum likelihood estimator for $\theta$ is defined as
 \begin{equation}
        \theta^* = \mathop{\arg\max}_{\theta} \prod_{i=1}^{n}p(x_i;\theta).
   \label{equ:mle}
 \end{equation}
This means finding the parameter that maximizes the joint density function of $n$ samples. 
For the convenience of calculation, the log-likelihood is often calculated, \emph{i.e.},
 \begin{equation}
        \theta^* = \mathop{\arg\max}_{\theta} \sum_{i=1}^{n}\log p(x_i;\theta).
   \label{equ:log-mle}
 \end{equation}

Thus, the log-likelihood function is derived from Eq. \eqref{equ:P_GPD}.

 \begin{equation}
   \begin{aligned}
        L_{GPD}(\bm{y}^j;\mu_j,\sigma_j,\xi_j) &=-o_j\log\sigma_j \\ 
          &-(1+\frac{1}{\xi_j})\sum^{o_j}_{i=1}\log(1+\xi_j\frac{y^j_i-\mu_j}{\sigma_j}),
   \end{aligned}
   \label{equ:llh}
 \end{equation}
where $y^j_i \geqslant \mu_j$ and $1+\xi_j\frac{y^j_i-\mu_j}{\sigma_j}>0$. When $\xi=0$, GPD is the exponential distribution, and the log-likelihood function is
 \begin{equation}
        L_{GPD}(\bm{y}^j;\mu_j,\sigma_j)=-o_j\log\sigma_j -\sigma_j^{-1}\sum_{i=1}^{o_j}(y^j_i-\mu_j).
   \label{equ:llh}
 \end{equation}
The $\mu_j$, $\sigma_j$ and $\xi_j$ is the corresponding location, scale and shape parameters for the $j$-th cluster.
Unfortunately, the maximum log-likelihood has no analytical solution and can only be solved numerically. Due to the upper limit of the maximum number of iterations, the parameter estimation by MLE is fast and constant in time complexity.
The MLE has some good convergence properties in comparison to other estimates (Method of Moments or Probability Weighted Moments).

\subsection{Optimization from the perspective of EM algorithm}
\label{sec:4.5}
Our GPD $k$-means can be intuitively understood from the perspective of 
EM algorithm. We give some insight discussion and analysis in this section. Particularly, the iterative algorithm of the $k$-means clustering is an Expectation-Maximization (EM) algorithm~\cite{wu1983convergence}.
Accordingly, the GPD $k$-means can be intuitively re-formulated as an EM algorithm by using the probability distribution of GPD. 




We introduce the vanilla EM formulation for the maximum likelihood with latent variable as follows,
\begin{equation}
		\begin{aligned}
			& p(\bm{x},z|\Theta) \propto \\
			& \left\{
			\begin{array}{rl}
				\exp(-\|\bm{x}-\bm{\theta}_{z}\|_2^2), & {\|\bm{x}-\bm{\theta}_{z}\|}_2 = \min_k {\|\bm{x}-\bm{\theta}_{k}\|}_2,\\
				0, & {\|\bm{x}-\bm{\theta}_{z}\|}_2 > \min_k {\|\bm{x}-\bm{\theta}_{k}\|}_2,
			\end{array} \right. 
		\end{aligned}
	\end{equation}
where $z$ is latent variable. In our GPD $k$-means, we replace ${\|\bm{x}-\bm{\theta}_{z}\|}_2$ with
$P(\bm{x},\bm{\theta}_{z})$ in the condition, 
\begin{equation}
   \begin{aligned}
& p(\bm{x},z|\Theta) \propto \\
& \left\{
\begin{array}{rl}
\exp(-\|\bm{x}-\bm{\theta}_{z}\|_2^2), & P(\bm{x},\bm{\theta}_{z}) = \max_k P(\bm{x},\bm{\theta}_k),\\
0, & P(\bm{x},\bm{\theta}_{z}) < \max_k P(\bm{x},\bm{\theta}_k),
\end{array} \right. 
   \end{aligned}
   \label{equ:mle}
\end{equation}
where the covering probability function in Eq. \ref{equ:P_GPD}.
When $P(\bm{x},\bm{\theta}_{z})$ between $\bm{x}$ and $\bm{\theta}_z$ is the largest, the probability is proportional to $\exp(-\|\bm{x}-\bm{\theta}_{z}\|_2^2)$, otherwise, it is 0.
In particular, $P(\bm{x},\bm{\theta})$  has three parameters estimated by MLE during the iteration. The estimated parameters will be used in the computation of covering probability in Eq.~\ref{equ:mle}.


In the EM algorithm, we optimize the following Q function via E step and M step.
Let $\bm{\Theta}^{(t)}$ be the estimated value of parameter $\bm{\Theta}$ in the $i$-th iteration. 
\begin{equation}
   \begin{aligned}
        Q(\bm{\Theta}, \bm{\Theta}^{(t)}) &= E_Z[\log p(X,Z|\bm{\Theta})|X,\bm{\Theta}^{(t)}] \\
        &= \sum_{i} \log p(\bm{x}_i,z_i=y_i|\bm{\Theta})\\
        &= const-\sum_{i} \|\bm{x}_i-\bm{\theta}_{y_i}\|_2^2.
   \end{aligned}
   \label{equ:Q}
\end{equation}

\noindent\textbf{E step} Calculating the following formula:

\begin{equation}
   \begin{aligned}
        p(z_i|x_i,\Theta^{(t)}) \propto \left\{
\begin{array}{rl}
1, & P(\bm{x}_i,\bm{\theta}_{z_i}^{(t)}) = \max_k P(\bm{x}_i,\bm{\theta}_k^{(t)}),\\
0, & P(\bm{x}_i,\bm{\theta}_{z_i}^{(t)}) < \max_k P(\bm{x}_i,\bm{\theta}_k^{(t)}).
\end{array} \right. 
   \end{aligned}
   \label{equ:E_step}
\end{equation}
This is equivalent to assigning each sample to the cluster with the maximum covering probability in GPD $k$-means.

\noindent\textbf{M step} Find the $\bm{\Theta}$ that maximizes the $Q(\bm{\Theta}, \bm{\Theta}^{(t)})$ as the estimated value of the parameter for the (t+1)-th iteration.
\begin{equation}
    \Theta^{(t+1)}=\mathop{\arg\max}_{\Theta}Q(\bm{\Theta}, \bm{\Theta}^{(t)}).
   \label{equ:max_Q}
\end{equation}
This is equivalent to $\Theta^{(t+1)}=\mathop{\arg\min}_{\Theta}\sum_{i} \|\bm{x}_i-\bm{\theta}_{y_i}\|_2^2$. At this time, the cluster label of each sample and parameters of GPD are determined, so the best centroids are equal to the average of all samples in each cluster.
 Therefore, the M step is equivalent to updating each centroid by averaging over its assigned samples. Therefore, GPD $k$-means will iteratively calculate E steps and M steps until convergence.

\begin{algorithm}[t]
\caption{GPD $k$-means}              
\label{alg:GPD_kmeans}
{\small
\begin{algorithmic}
  \STATE {\bfseries Input:} $\mathcal{X} \subseteq \mathbb{R}^{d}$, number of cluster $k$, threshold $u_j$.
  \STATE {\bfseries Output:} centroid $\Theta$
  \STATE {\bfseries Initialization:} random centroid $\Theta$
  \REPEAT
    \STATE Compute the Euclidean distance $d_{ij}^o$ and obtain $\{\bm{G}_1, \bm{G}_2, \ldots, \bm{G}_k\}$;
    \FOR{$j=1,2,\dots,k$}          
    \STATE Obtain $\bm{y}^j$ by Eq. \eqref{equ:y};
    \STATE Estimate the parameters of GPD on $\bm{y}^j$ by MLE;
    \ENDFOR
    
    \FOR{$i=1,2,\dots,n$}          
    \STATE $\lambda_i = \arg\max_{j \in \{1,2,\dots, k\}}P_{ij}$;
    \ENDFOR
    
    \FOR{$j=1,2,\dots,k$}          
    \STATE $\bm{\theta}_j = \frac{1}{|C_j|} \sum_{\bm{x} \in C_j}\bm{x}$;
    \ENDFOR
    
  \UNTIL{centroids no longer change}
        \STATE {\bfseries return} centroid $\Theta$\;  
\end{algorithmic} }
\end{algorithm}

As shown in Alg. \ref{alg:GPD_kmeans}, even though GPD $k$-means has one more step to fit GPD parameters with MLE than $k$-means, this step is only calculated on a small number of $\bm{y}^j$. As described in Sec. \ref{sec:4.4},  the time complexity of MLE is a constant, so the time complexity of GPD $k$-means is $O(nkd)$ like $k$-means.
The main difference between GPD k-means and k-means algorithm is that GPD k-means compares the probabilitis of the distances, while k-means algorithm compares the quantities of the distances. Therefore, 
our GPD k-means demand additional computational cost in fitting the distribution by maximum likelihood method. This is the only extra computational cost of our GPD k-means over k-means algorithm. This cost of computation time for a probalistic method like GPD k-means is essential as its start point is to learn a distribution model.
Regarding space complexity, GPD $k$-means uses $3k$ and $k$ more storage space  than $k$-means to store GPD parameters and log-likelihood, so the space complexity of GPD $k$-means is $O(nk)$ the same as $k$-means.

\section{Experiments and Results}
We evaluate clustering algorithms by four widely used metrics, unsupervised clustering accuracy (ACC) \cite{cai2010locally}, normalized mutual information (NMI) \cite{vinh2010information}, adjusted rand index (ARI) \cite{vinh2010information}, and Silhouette \cite{rousseeuw1987silhouettes}.
Note that the values of ACC and NMI are in the range of 0 to 1, with 1 indicating the best clustering and 0 indicating the worst clustering.
The values of ARI and Silhouette are in the range of -1 to 1, -1 indicates the worst clustering, and 1 indicates the best clustering.
There is no standard method for setting the two hyperparameters, the block size $s$ and the threshold $u$, so we set the $s$ by grid search and set $u$ adaptively. Specifically, we first set the hyperparameter $\alpha$ to indicate \emph{the percentage of excess} for all samples. 
Then we sort $-d_{ij}$ and set $u_j$ to the $\alpha$-th upper percentile of the sorted $-d_{ij}$. Furthermore, we set the percentage of excess $\alpha$ is 0.2.

\subsection{Synthetic Dataset Experiment}
\label{sec:5.1}
Our algorithm is compared to other algorithms on synthetic datasets we generated. To generate synthetic datasets, we select the relevant parameters, the number of samples $n$, the number of clusters $k$, and the dimension $d$ of the samples. We first uniformly sample $k$ centroids $\Theta$ in the hypercube $[-1,1]^d$. Then we generate $n/k$ samples by sampling $d$-dimensional vectors from a Gaussian distribution $N(0,\sigma^2)$. Finally, we add these Gaussian samples to the corresponding cluster centroid. 
We generated three synthetic datasets according to the steps described above. Their cluster numbers are $k=3, 4, 5$. The sample size $n$ is $1000$, the sample dimension $d$ is 2, and the $\sigma$ of Gaussian distribution is 0.2. 

\begin{figure*}[]
\centering
    \subfigure[Synthetic datasets ($n=1,000, k=3, d=2, \sigma=0.2$).]{
        \includegraphics[width=1.0\textwidth]{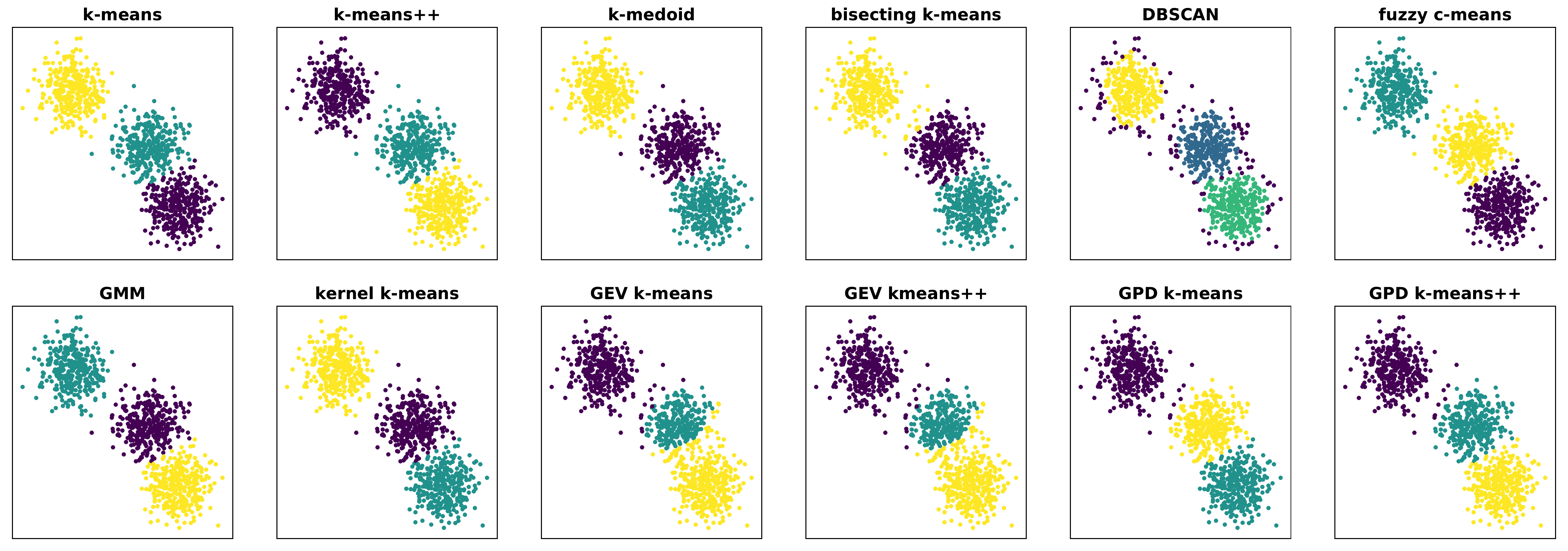}
        \label{fig:gaussian_k_3}
    }
    \subfigure[Synthetic datasets ($n=1,000, k=4, d=2, \sigma=0.2$).]{
        \includegraphics[width=1.0\textwidth]{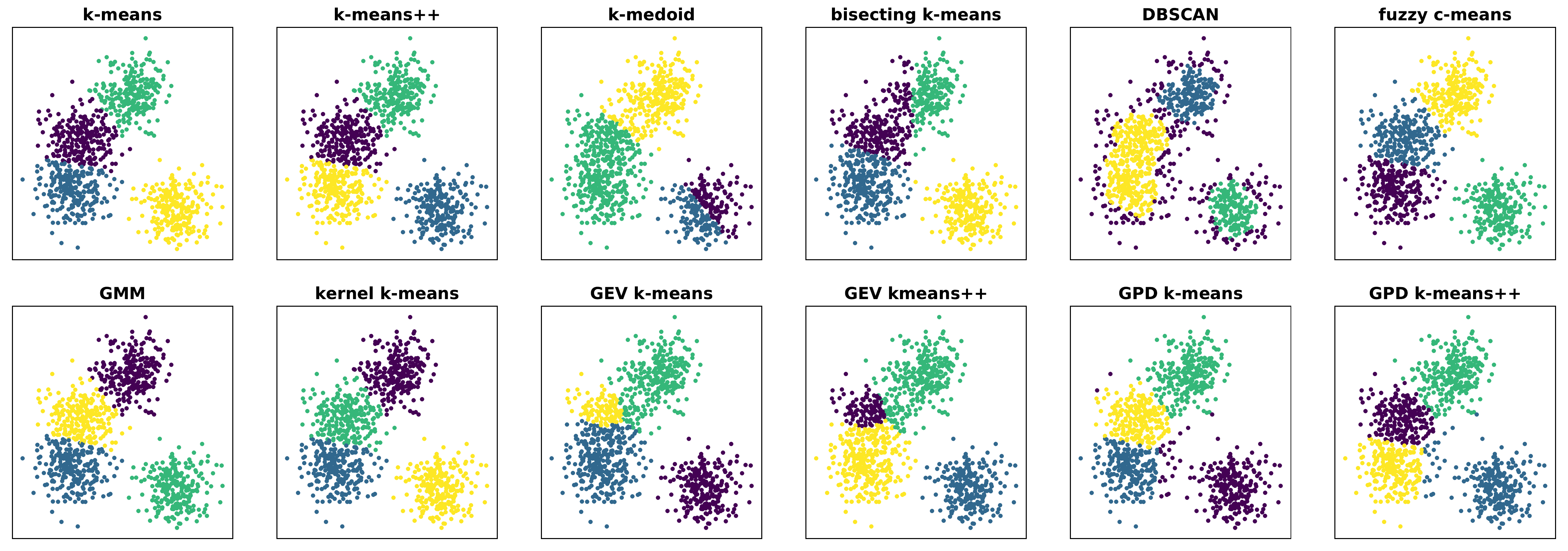}
        \label{fig:gaussian_k_4}
    }
    \subfigure[Synthetic datasets ($n=1,000, k=5, d=2, \sigma=0.2$).]{
        \includegraphics[width=1.0\textwidth]{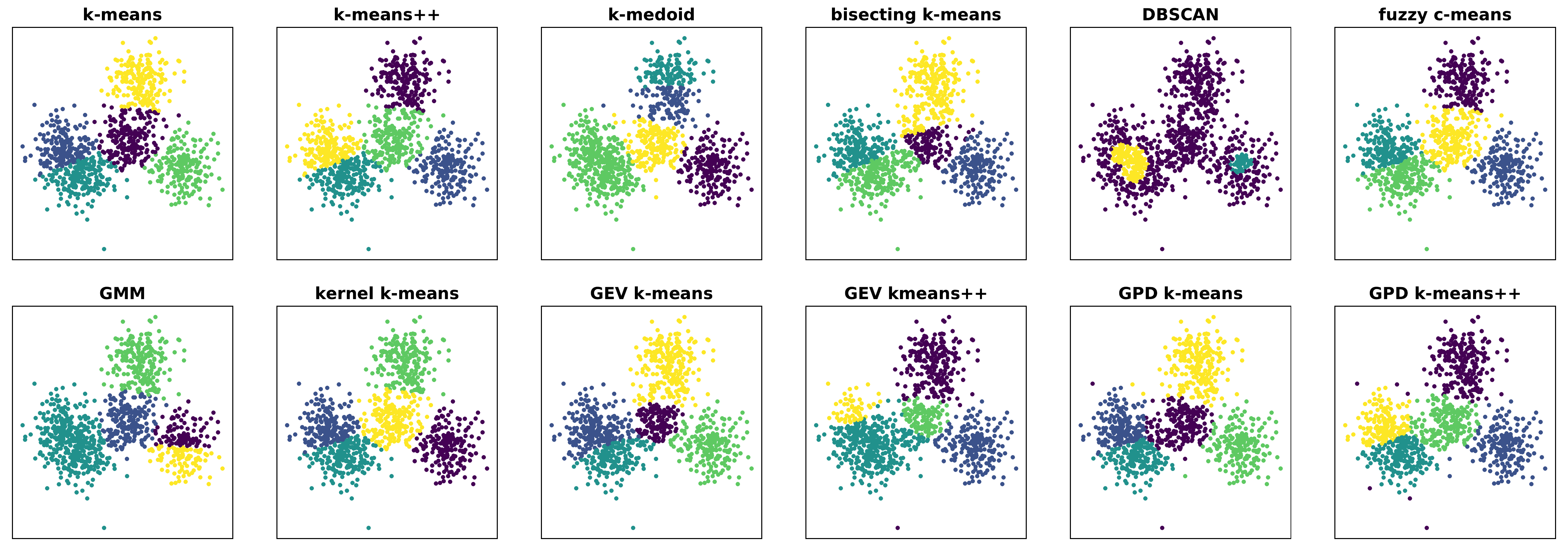}
        \label{fig:gaussian_k_5}
    }
    \caption{Visualization of synthetic datasets shows the result of our Extreme Value $k$-means compared to $k$-means, $k$-means++, $k$-medoid, bidecting $k$-means, DBSCAN, fuzzy c-means, GMM and kernel $k$-means.}
    \label{fig:gaussian}
\end{figure*}

As shown in Fig. \ref{fig:gaussian_k_3}, when $k=3$, the clustering results of the four variants of our algorithm (GEV $k$-means, GEV $k$-means++, GPD $k$-means, GPD $k$-means++) are basically the same. In addition, the clustering results of our GEV k-means and GPD $k$-means are similar to those of other algorithms. It is worth noting that DBSCAN regards low-density points far from clusters as noise points, and marks noise points as other colors.

As shown in Fig. \ref{fig:gaussian_k_4}, when the number of clusters reaches 4, the samples from different clusters touch and overlap each other. From the clustering results of our  GEV k-means and GPD $k$-means, we can see that the decision boundary of  GEV k-means and GPD $k$-means is not a straight line, but a curve. This is because the decision boundary of  GEV k-means and GPD $k$-means is a contour with zero covering probability. The clustering result of our GEV k-means and GPD $k$-means is not much different from other algorithms (except DBSCAN).

As shown in Fig. \ref{fig:gaussian_k_5}, when the number of clusters reaches 5, the samples of different clusters touch and overlap each other more seriously. GMM clusters the samples in the lower-right corner into two cluster instead of one clusters. $k$-medoid clusters the samples in the lower-left corner into one cluster instead of two clusters. The clustering result of DBSCAN is not good. On the contrary, our  GEV k-means and GPD $k$-means get better clustering results, and it is closer to the clustering results of $k$-means and $k$-means++.

\begin{figure*}[]
  \centering
      \subfigure[\small{The impact of different sample size $n$ on ARI}]{
         \includegraphics[width=0.45\linewidth]{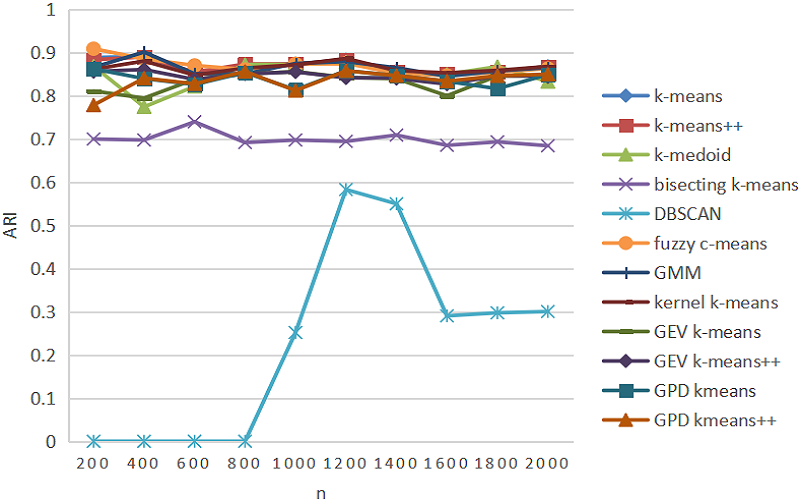}
         \label{fig:batch_size}
      }
      \subfigure[\small{The impact of different sample dimension $d$ on ARI}]{
         \includegraphics[width=0.45\linewidth]{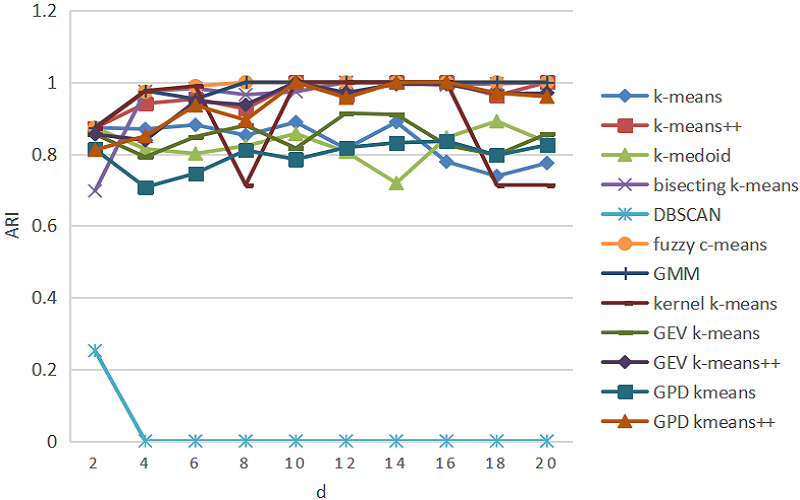}
         \label{fig:batch_size++}
      }
        \subfigure[\small{The impact of different cluster numbers $k$ on ARI}]{
         \includegraphics[width=0.45\linewidth]{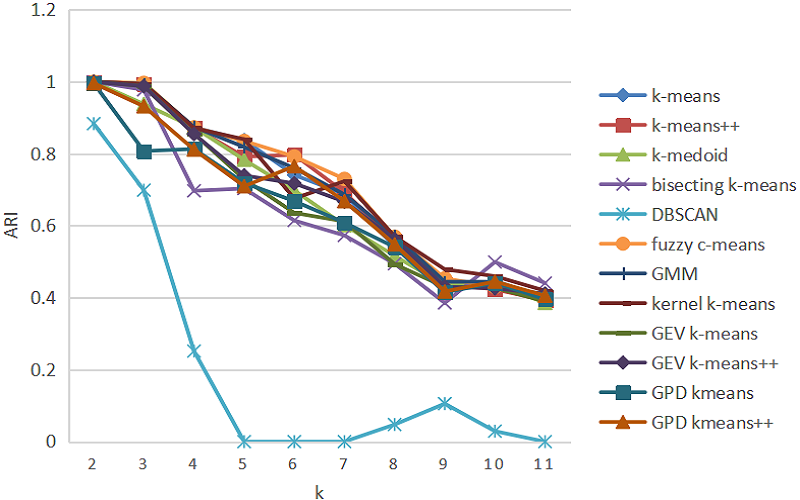}
         \label{fig:batch_size}
      }
      \subfigure[\small{The impact of different the std of Gaussian distribution on ARI}]{
         \includegraphics[width=0.45\linewidth]{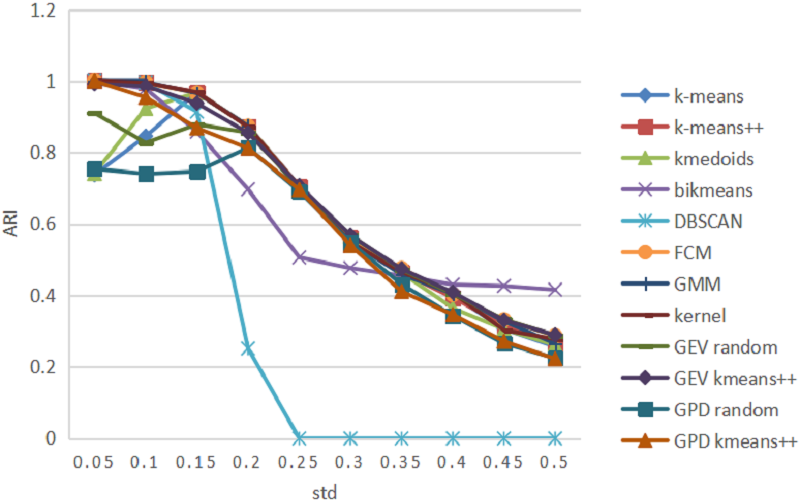}
         \label{fig:batch_size++}
      }
  \caption{Ablation study on synthetic dataset.}
  \label{fig:ablating}
\end{figure*}

As shown in Fig. \ref{fig:ablating}, we studied the effect of the sample size $n$, sample dimension $d$, cluster number $k$, and the standard deviation (std) of Gaussian distribution on the ARI on the synthetic dataset. 
We found that except for DBSCAN, the trend of other algorithms is basically the same. This may be because the hyperparameters of DBSCAN are not very adaptable.
And we found that $n$ and $d$ have little effect on ARI. As $k$ and std increase, ARI gradually decreases.

\begin{table*}
	\caption{Results of GEV $k$-means, GPD $k$-means and other algorithms on nine real datasets.}
	\label{res:ev_$k$-means}
	\centering
	\setlength{\tabcolsep}{0.4em}
	\small\footnotesize
	\begin{tabular}{l|cccc|cccc}
		\hline\hline
		\multirow{2}{*}{Algorithm} & ACC             & ARI             & NMI             & Silhouette      & ACC             & ARI             & NMI             & Silhouette      \\ \cline{2-9} 
		& \multicolumn{4}{c|}{sonar}                                            & \multicolumn{4}{c}{heart}                                                                                      \\ \hline\hline
		$k$-means                    & 0.5490          & 0.0065          & 0.0123          & 0.1689          & 0.8156          & 0.3964          & 0.3084          & 0.1688                    \\
		$k$-means++                  & 0.5423          & 0.0033          & 0.0139          & 0.2048          & 0.8207          & 0.4096          & 0.3232          & 0.1704                    \\
		$k$-medoids                  & 0.5625          & 0.0180          & 0.0251          & 0.1399          & 0.7996          & 0.3591          & 0.2865          & 0.1483                    \\
		bisecting $k$-means          & 0.5591          & 0.0140          & 0.0196          & 0.1826          & 0.8156          & 0.3964          & 0.3084          & 0.1708                   \\
		DBSCAN                     & 0.5104          & -0.0038         & 0.0294          & 0.0000          & 0.6185          & 0.0329          & 0.0213          & 0.0000                    \\
		FCM                        & 0.5707          & 0.0198          & 0.0182          & 0.1338          & 0.8048          & 0.3694          & 0.2862          & 0.1604                    \\
		GMM                        & 0.5486          & 0.0068          & 0.0135          & 0.1867          & 0.7985          & 0.3550          & 0.2714          & 0.1499                    \\
		kernel $k$-means             & 0.5962          & 0.0323          & 0.0252          & 0.1163          & 0.7556          & 0.2583          & 0.2434          & 0.1352                    \\
		GEV $k$-means                & 0.6010          & 0.0363          & 0.0448          & 0.1809          & 0.8289          & 0.4307          & 0.5489          & 0.1443                    \\
		GEV $k$-means++              & \textbf{0.6202} & \textbf{0.0533} & \textbf{0.0577} & \textbf{0.2861} & 0.8289          & 0.4307          & \textbf{0.9509} & 0.1443                    \\
		GPD $k$-means                & 0.5865          & 0.0254          & 0.0294          & 0.2218          & 0.8441          & 0.4716          & 0.3639          & 0.1759                    \\
		GPD $k$-means++              & 0.6154          & 0.0487          & 0.0490          & 0.2218          & \textbf{0.8463} & \textbf{0.4775} & 0.3472          & \textbf{0.1760}  \\ \hline\hline
		& \multicolumn{4}{c|}{vehicle}                                           & \multicolumn{4}{c}{fourclass} \\ \hline\hline
		$k$-means                    & 0.3687          & 0.0816          & 0.1200          & 0.2678     & 0.6439          & 0.0813          & 0.0577          & 0.3491          \\
		$k$-means++                  & 0.3636          & 0.0751          & 0.1120          & 0.2653     & 0.6531          & 0.0922          & 0.0644          & 0.3420          \\
		$k$-medoids                  & 0.3635          & 0.0767          & 0.1152          & 0.2430     & 0.6297          & 0.0907          & 0.0861          & 0.3376           \\
		bisecting $k$-means          & 0.2839          & 0.0024          & 0.0069          & -0.0258    & 0.6672          & 0.1147          & 0.0873          & 0.3515                    \\
		DBSCAN                     & 0.2611          & 0.0007          & 0.0172          & 0.0000    & 0.6439          & 0.0000          & 0.0000          & 0.0000                    \\
		FCM                        & 0.3700          & 0.0711          & 0.0875          & 0.2295    & 0.6485          & 0.0869          & 0.0642          & 0.3571                    \\
		GMM                        & 0.3826          & 0.0942          & 0.1411          & 0.2483    & 0.6653          & 0.1172          & 0.0874          & 0.3498          \\
		kernel $k$-means             & 0.3641          & 0.0618          & 0.0880          & 0.2115     & 0.6334          & 0.0634          & 0.0299          & 0.3221          \\
		GEV $k$-means                & 0.3933          & 0.0924          & 0.1291          & 0.2600     & 0.6833          & 0.1566          & 0.1082          & 0.3339          \\
		GEV $k$-means++              & 0.3632          & 0.0809          & 0.1308          & 0.2020     & 0.7156          & 0.1834          & 0.1245          & 0.3339          \\
		GPD $k$-means                & 0.3803          & \textbf{0.0998} & 0.1304          & 0.2686     & 0.7390          & 0.2265          & 0.1651          & 0.3572          \\
		GPD $k$-means++              & \textbf{0.3940} & 0.0991          & \textbf{0.1466} & \textbf{0.2694}     & \textbf{0.7425} & \textbf{0.2327} & \textbf{0.1781} & \textbf{0.3573} \\ \hline\hline
      & \multicolumn{4}{c|}{poker}                                            & \multicolumn{4}{c}{cod-rna} \\ \hline\hline
      $k$-means                     & 0.1085          & 0.0003          & 0.0020          & 0.0677          & 0.5509          & -0.0169         & 0.0048          & 0.2871          \\
      $k$-means++                   & 0.1086          & 0.0002          & 0.0018          & 0.0676          & 0.5528          & -0.0169         & 0.0047          & 0.2876          \\
      $k$-medoids                   & 0.1413          & 0.0002          & 0.0021          & 0.0497          & 0.5134          & 0.0009          & 0.0009          & 0.2262         \\
      bisecting $k$-means           & 0.1300          & 0.0001          & 0.0013          & -0.0069         & 0.5512          & -0.0169         & 0.0048          & 0.2860\\
      DBSCAN                        & 0.1101          & 0.0000          & 0.0017          & 0.0000          & 0.6667          & 0.0000          & 0.0000          & 0.0000\\
      FCM                           & 0.1732          & 0.0001          & 0.0019          & 0.0246          & 0.5280          & 0.0030          & 0.0039          & 0.2231\\
      GMM                           & 0.2054          & 0.0001          & 0.0019          & -0.0150         & 0.5016          & 0.0000          & 0.0000          & 0.2269          \\
      kernel $k$-means              & 0.1087          & 0.0001          & 0.0014          & 0.0682          & 0.5124          & -0.0002         & 0.0000          & 0.2068          \\
      GEV $k$-means                 & 0.4950          & 0.0026          & 0.0027          & 0.6870          & \textbf{0.6938} & \textbf{0.0680} & \textbf{0.0597} & \textbf{0.3156}  \\
      GEV $k$-means++               & \textbf{0.5005} & \textbf{0.0042} & 0.0023          & \textbf{0.6890} & 0.6713          & 0.0130          & 0.0597          & 0.3052          \\
      GPD $k$-means                 & 0.2495          & 0.0011          & \textbf{0.0041} & 0.0679          & 0.6656          & 0.0066          & 0.0184          & 0.3012          \\
      GPD $k$-means++               & 0.2497          & 0.0008          & 0.0039          & 0.0681          & 0.6646          & 0.0163          & 0.0337          & 0.3022          \\ \hline\hline
		& \multicolumn{4}{c|}{usps}                                             & \multicolumn{4}{c}{MNIST feature}                                           \\ \hline\hline
		$k$-means                     & 0.5434          & 0.4375          & 0.5462          & 0.1461          & 0.8250          & 0.8122          & 0.8879          & 0.3059                    \\
		$k$-means++                   & 0.5451          & 0.4286          & 0.5456          & 0.1465          & 0.8482          & 0.8286          & 0.8930          & 0.3390                    \\
		$k$-medoids                   & 0.5143          & 0.3810          & 0.4967          & 0.0913          & 0.7703          & 0.7286          & 0.8346          & 0.2564                    \\
		bisecting $k$-means           & 0.1380          & 0.0042          & 0.0087          & -0.0175         & 0.1050          & 0.0000          & 0.0003          & -0.0056                  \\
		DBSCAN                        & 0.0000          & 0.0825          & 0.3361          & -0.1635         & 0.0000          & 0.0000          & 0.5000          & 0.0000                    \\
		FCM                           & 0.2965          & 0.1112          & 0.1820          & -0.0193         & 0.4668          & 0.3791          & 0.5881          & 0.0120                    \\
		GMM                           & 0.5197          & 0.3860          & 0.5138          & 0.1246          & 0.8823          & 0.7642          & 0.8313          & 0.2996                    \\
		kernel $k$-means              & 0.6386          & 0.4787          & 0.5700          & 0.1128          & 0.8634          & 0.8099          & 0.8760          & 0.3079                    \\
		GEV $k$-means                 & 0.5796          & 0.3777          & 0.5049          & 0.1422          & 0.7479          & 0.6447          & 0.7469          & 0.3145                    \\
		GEV $k$-means++               & 0.5471          & 0.3309          & 0.4911          & 0.1408          & \textbf{0.9330} & 0.8616          & 0.8685          & 0.3277                   \\
		GPD $k$-means                 & \textbf{0.6456} & \textbf{0.4943} & \textbf{0.6022} & \textbf{0.1577} & 0.8210          & 0.7989          & 0.8678          & 0.3509                   \\
		GPD $k$-means++               & 0.6125          & 0.4785          & 0.5603          & 0.1365          & 0.9201          & \textbf{0.8965} & \textbf{0.9147} & \textbf{0.3832} \\
		\hline\hline
                                    & \multicolumn{4}{c|}{CIFAR10}                                           & \multicolumn{4}{c}{MNIST raw}                                           \\ \hline\hline
      $k$-means                     & 0.7512          & 0.7519          & 0.9123          & 0.6172          & 0.4774 & 0.3066 & 0.4211 & 0.0066  \\
      $k$-means++                   & 0.8256          & 0.8304          & 0.9179          & 0.8999          & 0.4658 & 0.2929 & 0.4089 & 0.0022  \\
      $k$-medoids                   & 0.7847          & 0.7640          & 0.9199          & 0.6482          & 0.3783 & 0.1831 & 0.3055 & \textbf{0.0105}  \\
      bisecting $k$-means           & 0.1054          & 0.0000          & 0.0003          & -0.0055         & 0.1060 & 0.0000 & 0.0003 & -0.0156 \\
      DBSCAN                        & 0.0000          & 0.0044          & 0.1970          & 0.8229          & 0.0000 & 0.0000 & 0.0000 & 0.0000  \\
      FCM                           & 0.3868          & 0.3672          & 0.6860          & 0.0982          & 0.2776 & 0.0931 & 0.1904 & -0.0216 \\
      GMM                           & 0.9775          & 0.9634          & 0.9387          & 0.8999          & 0.4007 & 0.1924 & 0.3510 & -0.0628 \\
      kernel $k$-means              & 0.9000          & 0.8979          & 0.9690          & 0.8052          & \textbf{0.5038} & \textbf{0.3159} & \textbf{0.4294} & -0.0123 \\
      GEV $k$-means                 & 0.8397          & 0.8318          & 0.9466          & 0.7741          & 0.1637 & 0.0099 & 0.1381 & -0.2008 \\
      GEV $k$-means++               & 0.9801          & 0.9737          & 0.9924          & 0.8704          & 0.1650 & 0.0079 & 0.1651 & -0.1867 \\
      GPD $k$-means                 & 0.8595          & 0.8465          & 0.9497          & 0.7862          & 0.3472 & 0.1074 & 0.3407 & -0.0776 \\
      GPD $k$-means++               & \textbf{0.9881} & \textbf{0.9860} & \textbf{0.9932} & \textbf{0.9256} & 0.3656 & 0.1137 & 0.3437 & -0.0764 \\           
      \hline\hline                        
	\end{tabular}
\end{table*}

\begin{table*}
	\caption{Results of the time costs of our algorithms and vanilla k-means on three real datasets.}
	\label{res:time}
	\centering
	\setlength{\tabcolsep}{0.4em}
	\small\footnotesize
	\begin{tabular}{l|rrrrr}
		\hline
		method         & total time & total MLE time & avg MLE time & total cluster time & avg cluster time \\\hline\hline
		        &\multicolumn{5}{c}{cod-rna}\\\hline
        GEV k-means           & 11.267      & 9.442            & 0.111          & 1.825                & 0.021              \\
        GEV k-means++          & 12.352      & 10.369           & 0.109          & 1.982                & 0.021              \\
        GPD k-means           & 2.592       & 2.393            & 0.263          & 0.199                & 0.022              \\
        GPD k-means++          & 3.198       & 2.948            & 0.256          & 0.249                & 0.022              \\
        k-means         & 0.205       & 0.000            & 0.000          & 0.205                & 0.007              \\
        k-means++       & 0.220       & 0.000            & 0.000          & 0.220                & 0.007              \\\hline\hline
		 & \multicolumn{5}{c}{MNIST raw} \\\hline
        GEV k-means           & 94.746      & 40.390           & 0.404          & 54.356               & 0.544              \\
        GEV k-means++          & 95.585      & 41.231           & 0.412          & 54.355               & 0.544              \\
        GPD k-means            & 236.684     & 181.253          & 1.813          & 55.431               & 0.554              \\
        GPD k-means++          & 242.714     & 187.450          & 1.874          & 55.264               & 0.553              \\
        k-means         & 38.769      & 0.000            & 0.000          & 38.769               & 0.475              \\
        k-means++       & 46.379      & 0.000            & 0.000          & 46.379               & 0.479              \\\hline\hline
		        & \multicolumn{5}{c}{CIFAR10} \\\hline
        GEV k-means            & 71.160      & 40.051           & 0.401          & 31.109               & 0.311              \\
        GEV k-means++          & 9.803       & 5.782            & 0.464          & 4.021                & 0.360              \\
        GPD k-means            & 47.935      & 39.887           & 1.571          & 8.048                & 0.337              \\
        GPD k-means++          & 7.388       & 5.973            & 1.493          & 1.415                & 0.354              \\
        k-means         & 8.360       & 0.000            & 0.000          & 8.360                & 0.245              \\
        k-means++       & 0.506       & 0.000            & 0.000          & 0.506                & 0.241             \\ \hline
	\end{tabular}
\end{table*}

\subsection{Real Dataset Experiment} \label{sec:4.2}
We evaluate our GEV $k$-means and GPD $k$-means on nine real datasets: sonar ($n=208,d=60,k=2$), heart ($n=270,d=13,k=2$), vehicle ($n=846,d=18,k=4$),  fourclass ($n=862,d=2,k=2$), poker ($n=25,010,d=10,k=10$), cod-rna ($n=59,535,d=8,k=2$), usps ($n=7,291,d=256,k=10$), MNIST \cite{lecun1998mnist} ($n=60,000, d=84, k=10$) and CIFAR10 \cite{krizhevsky2009learning} ($n=50,000, d=512, k=10$). The first seven datasets are available from UCI repository \cite{asuncion2007uci} and LIBSVM Data website \cite{chang2011libsvm}.
MNIST is a dataset comprises 70,000 grey-scale images of handwritten digits 0 to 9. We examine our algorithms on two different versions of the MNIST dataset. The first is the original MNIST consisting of $28 \times 28$ grey-scale images, denoted as MNIST raw. The second is the 84-dimensional features obtained by LeNet\cite{lecun1998gradient}.
CIFAR10 is a dataset containing 60,000 color images with $32 \times 32$ pixels, grouped into 10 different classes of equal size, representing 10 different objects. Each of the training images is represented by a 512-dimensional vector extracted by a ResNet-18 \cite{he2016deep}. Each feature of all datasets is  normalized to unit variance.
We compare four variants (GEV $k$-means, GEV $k$-means++, GPD $k$-means, GPD $k$-means++) with eight other algorithms ($k$-means, $k$-means++, $k$-medoid \cite{kaufman2009finding}, bisecting $k$-means \cite{karypis2000comparison},  DBSCAN \cite{ester1996density}, fuzzy $c$-means \cite{dunn1973fuzzy,bezdek2013pattern}, GMM \cite{bishop2006pattern}, kernel $k$-means\cite{scholkopf1998nonlinear}). GEV $k$-means++ and GPD $k$-means++ indicate the use of $k$-means++ to initialize the centroids.
We repeat each experiment 10 times with different random seeds and take the mean of the results of 10 times experiments as the final result. 


\begin{figure*}[h]
  \centering
      \subfigure[\small{The fitting of GEV}]{
         \includegraphics[width=0.24\linewidth]{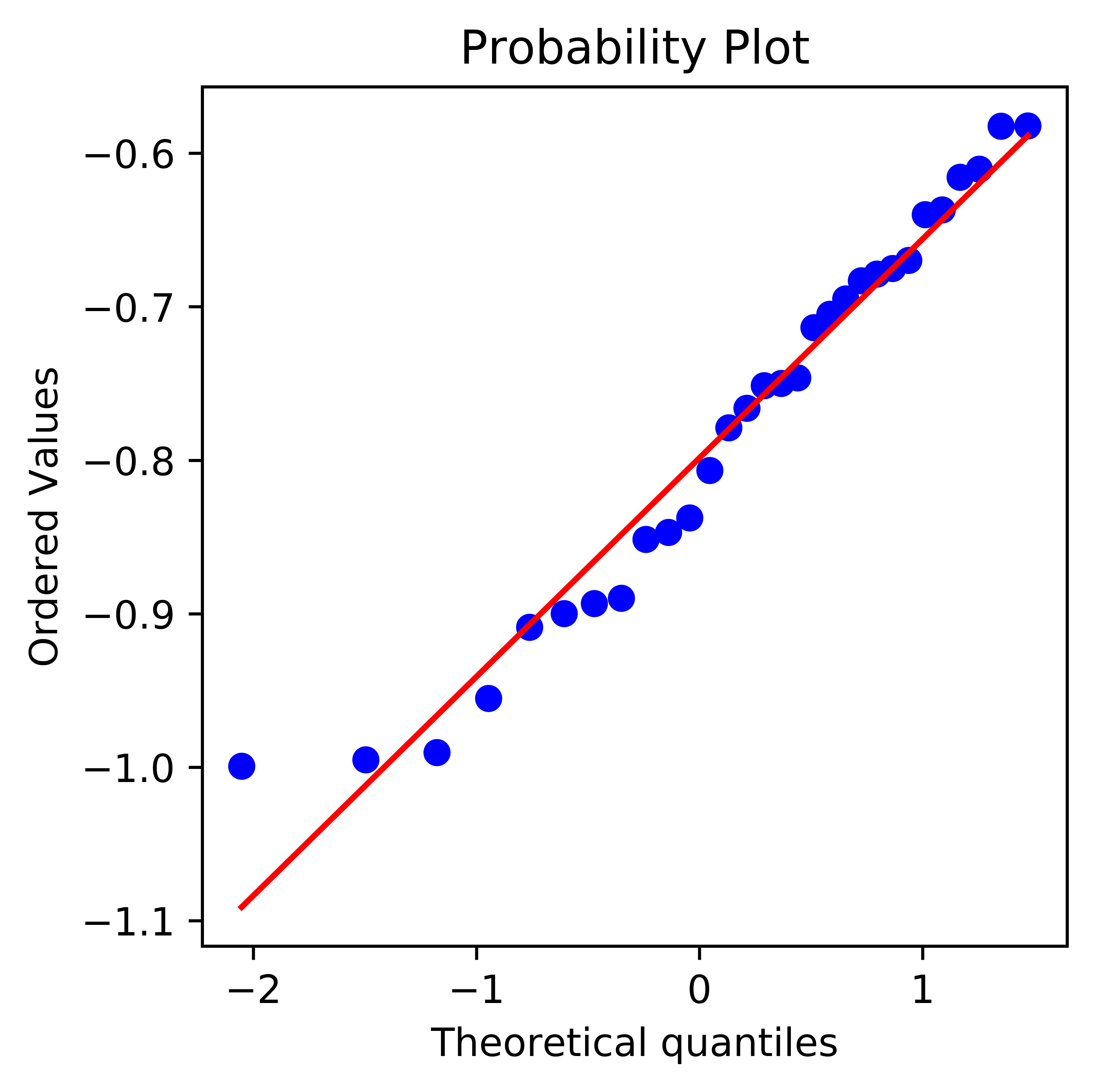}
         \label{fig:ablation_a}
         \hspace{-0.2in}
      }
      \subfigure[\small{The fitting of GPD}]{
         \includegraphics[width=0.25\linewidth]{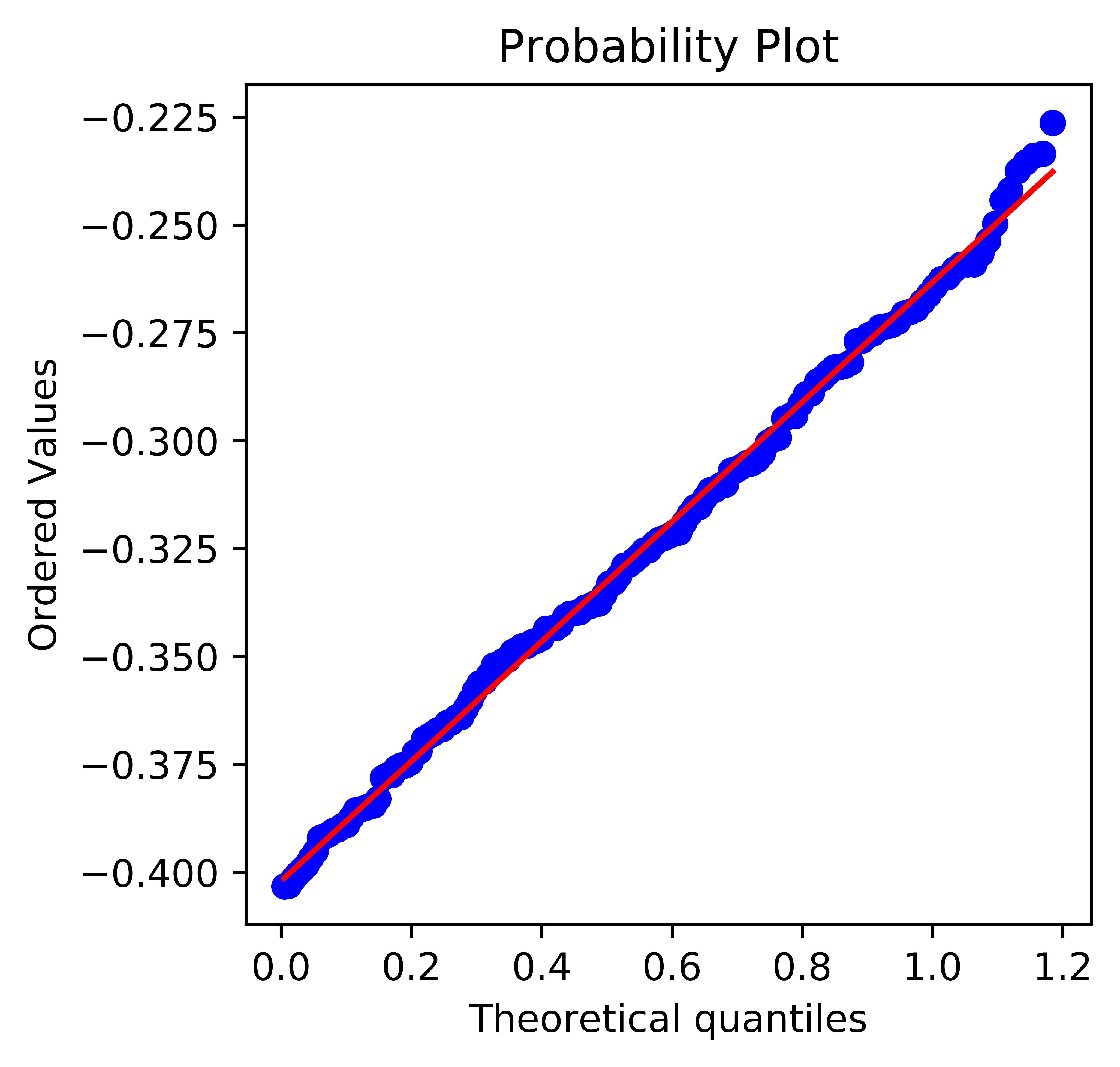}
         \label{fig:ablation_b}
         \hspace{-0.2in}
      }
      \subfigure[\small{Analysis of $s$}]{
         \includegraphics[width=0.24\linewidth]{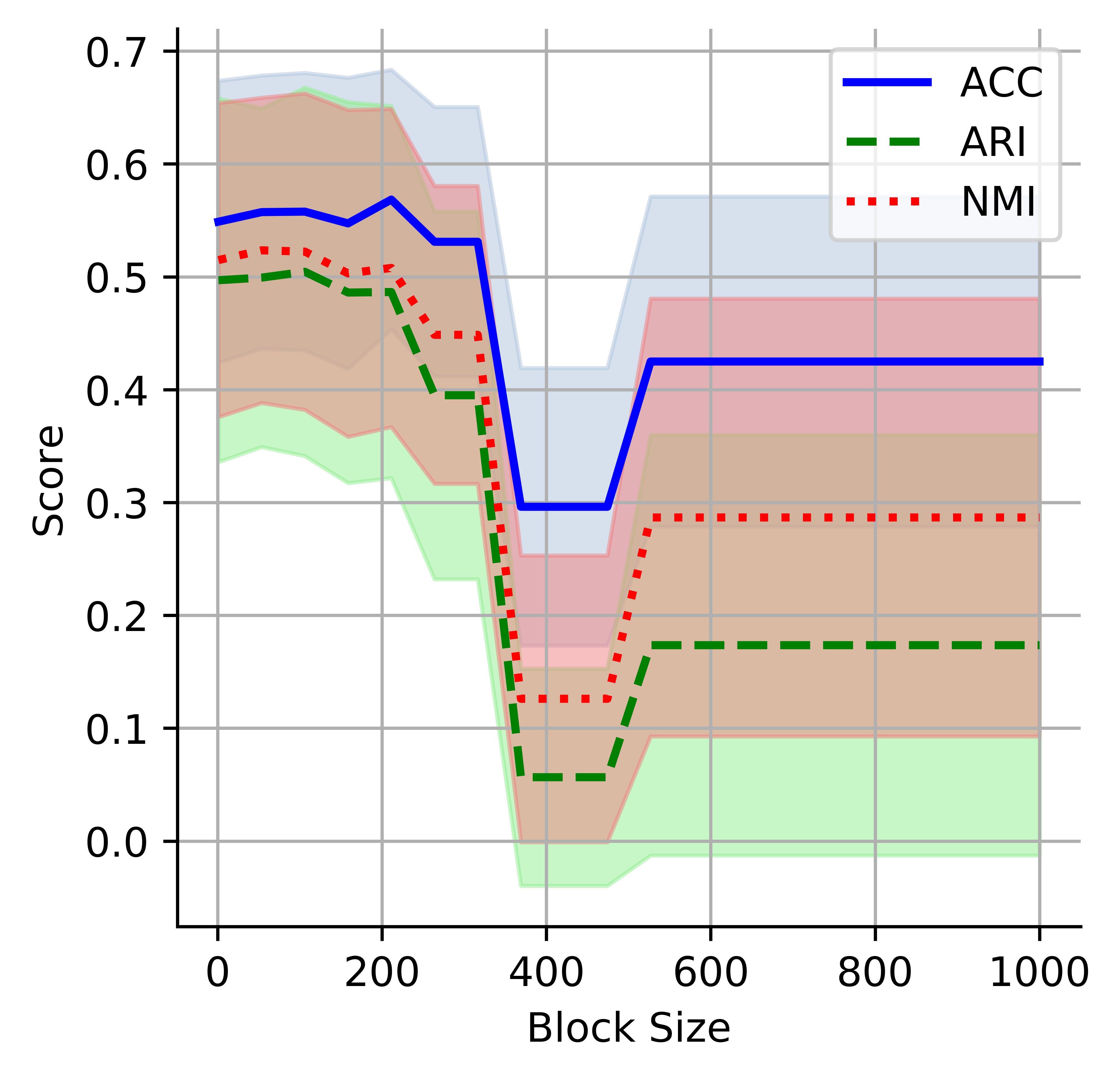}
         \label{fig:ablation_c}
         \hspace{-0.2in}
      }
      \subfigure[\small{Analysis of $\alpha$}]{
         \includegraphics[width=0.24\linewidth]{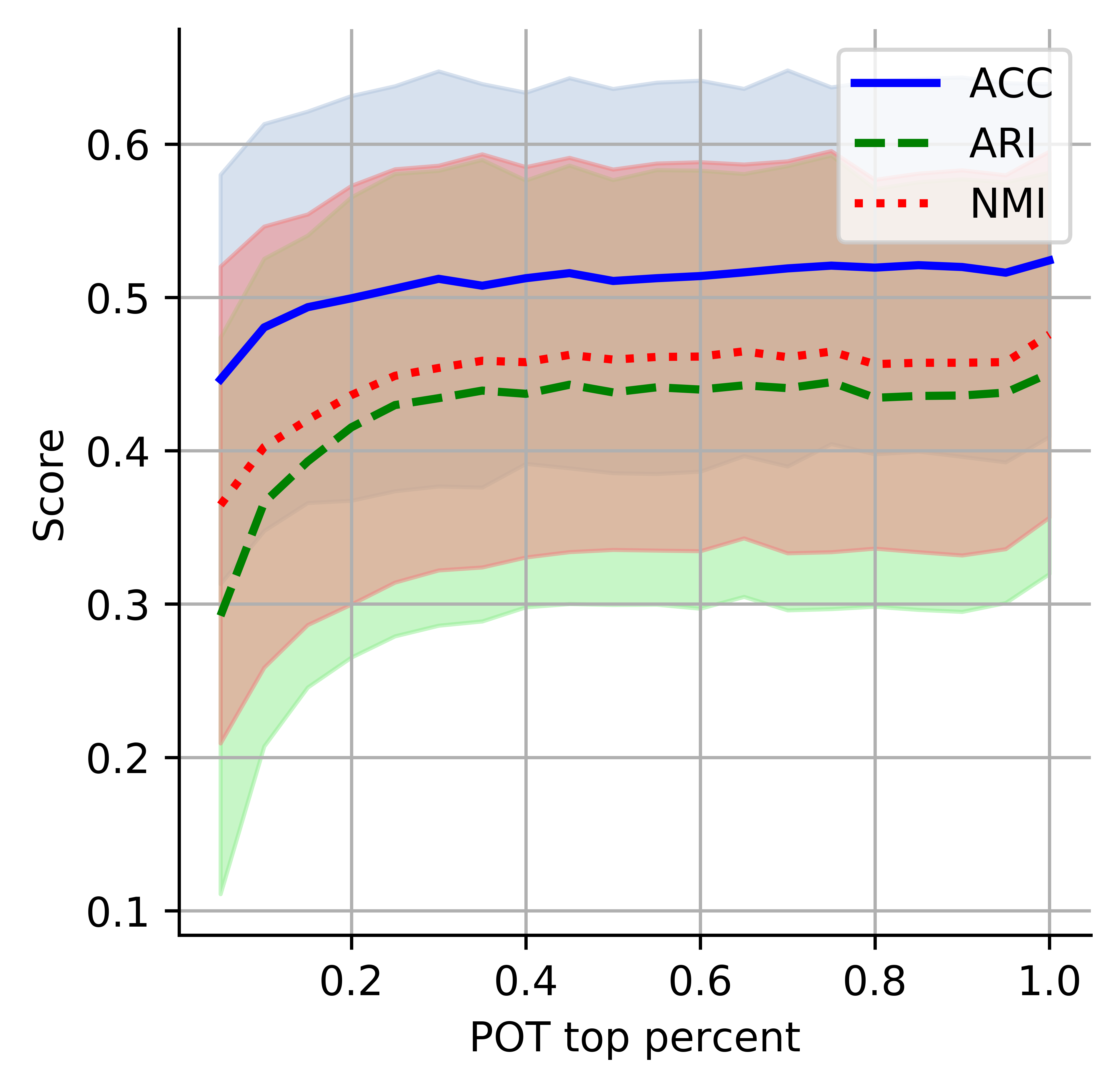}
         \label{fig:ablation_d}
      }
  \caption{Analysis of the fitting of GEV and GPD, block size $s$ and the percentage of excess $\alpha$ on a synthetic dataset  (Sec. \ref{sec:5.1}) with $n=1000, k=4, d=2, \sigma=0.3$.}
\end{figure*}

The clustering results on nine real datasets (including two versions of MNIST dataset) are shown in Tab. \ref{res:ev_$k$-means}. We can consider the four datasets of sonar, heart, vehicle, and fourclass as small dataset, because their data volume $n$ is less than 1000, and the other five data sets are considered as large datasets. 
As shown in Tab. \ref{res:ev_$k$-means}, our GEV $k$-means and GPD $k$-means outperform other algorithms on all nine datasets except the raw MNIST dataset. The experimental results show that most algorithms have better results on the features extracted from MNIST than the raw MNIST dataset thus, the MNIST feature is more suitable for clustering. In other words, it is better to apply deep learning techniques to boost clustering. It can be observed from Tab. \ref{res:ev_$k$-means} that our GPD $k$-means performs better on large datasets. For example, on the heart dataset, GPD $k$-means++ has an ACC score higher than $k$-means++ by 0.0256. On the CIFAR10 dataset, GPD $k$-means++ has an ACC score 0.1625 higher than $k$-means++. 
Secondly, we can obtain better performance by using $k$-means++ to initialize the centroids of GEV $k$-means or GPD $k$-means in the case of large number than random initializing the centroid.
The four variants of our algorithm perform similarly on nine datasets. On heart, vehicle, fourclass, usps, MNIST, and CIFAR10, the performance of GPD $k$-means and GPD $k$-means++ are better than GEV $k$-means and GEV $k$-means++.In addition, the performance of GPD $k$-means and GPD $k$-means++ is more stable.

\textbf{Time cost comparison} We examine the time costs of our algorithms and traditional k-means algorithms in Tab.\ref{res:time}. We repeat each experiment 10 times with different random seeds and take the mean as the final result. As for the codes of k-means and k-means++, we utilize the python realization of Llyod's algorithm.
Critically,  to make a fair comparison, 
we do not use the sklearn realization of k-means, since it  has many important acceleration tricks, such as OpenMP-based parallelism through Cython.  Specifically, we separate the computational cost of the MLE and the clustering stages of our algorithm. We have the following conclusions: 
First, the clustering time of our GPD k-mean is comparable to those of k-means and k-means++. Importantly,  as the core of our paper is a new cluster method, the pure computational time for clustering is not too costly.
Second, our algorithm demands significant computational time for the covering probabilities by MLE, while k-mean and k-means++ do not need such a stage. Thus our algorithm will take more time at this stage. Therefore, it is an important future work of accelerating the computation of MLE, which is beyond this paper's scope.

\subsection{Robustness to Uninformative Features}

\begin{figure*}
\centering
\includegraphics[width=\linewidth]{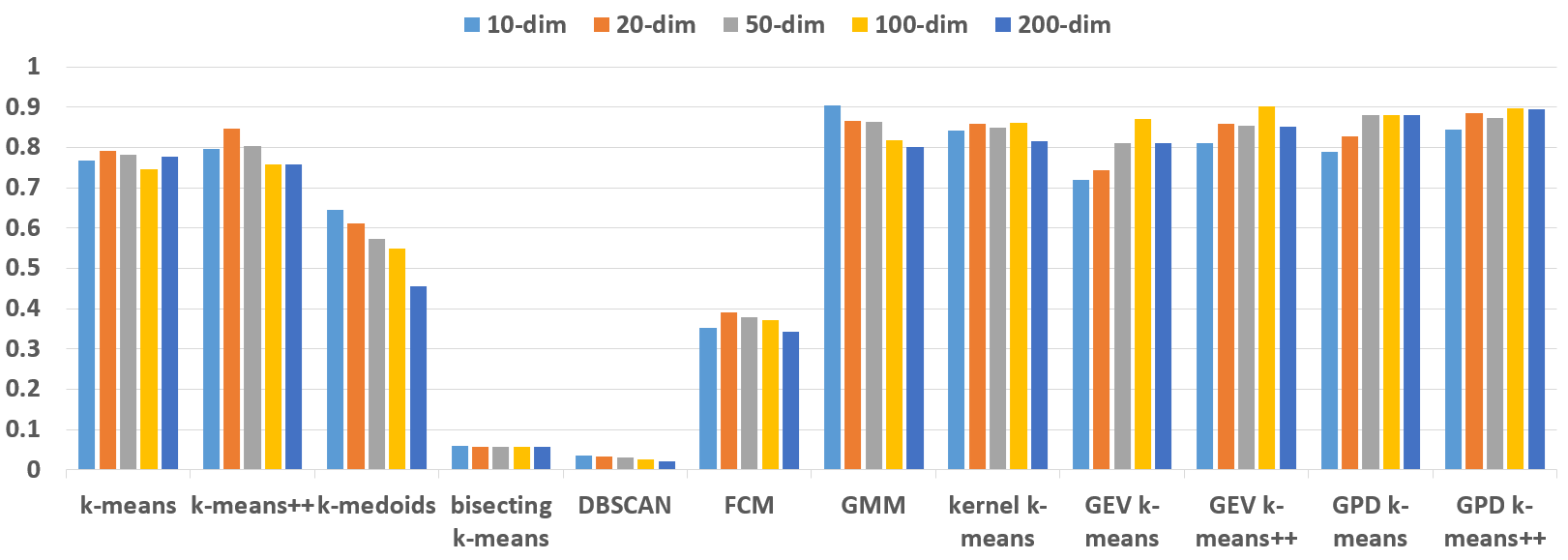}
\vspace{-0.2in}
\caption{The impact of increasing the number of uninformative features on ACC.}
\label{fig:add_dim}
\vspace{-0.2in}
\end{figure*}

In many cases, the dataset may contain uninformative features that affect the performance of the clustering algorithm. Therefore, we studied the influence of increasing number of uninformative features on the metric ACC. We first generate  Gaussian distribution samples with $n=4000, d=50, k=50$ and standard deviation $std=2$. Each sample is added 10-d, 20-d, 50-d, 100-d, and 200-d uninformative features, which are generated from the $N(0,1)$ Gaussian distribution. As shown in Fig. \ref{fig:add_dim}, the ACC of other algorithms decreases or changes little with the increase of the number of uninformative features. However, the ACC of the four variants of our algorithm does not decline, but has a slow upward trend, indicating that our algorithm is very robust to uninformative features.

\subsection{Analysis of GEV and GPD Fitting}

We conducted experiments on the synthetic data of the parameter $n=1000, k=4, d=2, \sigma=0.3$ and analyzed the fitting of GEV and GPD, the effect of the block size $s$ and the percentage of excess $\alpha$ on the performance.
To test the fitting of GEV and GPD, we used the most commonly used Quantile-Quantile (Q-Q) plot. Q-Q plot is a graphical technique for determining whether a certain two datasets are from the same distribution. As shown in Fig. \ref{fig:ablation_a} and \ref{fig:ablation_b}, we select a centroid fitting result and draw Q-Q plots, which are very approximate to a straight line, indicating that GEV and GPD fitting is very well.
In order to study the effect of different block size $s$ on GEV $k$-means and the percentage of excess $\alpha$ on GPD $k$-means, we used different $s$ and $\alpha$ to perform experiments. 
Fig. \ref{fig:ablation_c} shows that as the block size increases, ACC, ARI, and NMI first appeared to rise, then fell sharply and maintained.
This confirms Theorem \ref{theorem:FT} that the block size should be large enough, but too large will cause too little extreme data and cause the fitting to fail.
As shown in Fig. \ref{fig:ablation_d}, as $\alpha$ increases, ACC, ARI, and NMI increase first and then maintain a slight fluctuation. Therefore, $\alpha$ should be a relatively small value to get a large enough $u$, because no increase in $\alpha$ can get a great performance improvement.

\noindent \textbf{Remark about i.i.d assumption.} In statistics and machine learning, it is commonly assumed that observations in a sample are effectively i.i.d, which can simplify the underlying mathematics of many statistical methods. Similarly, this paper takes the i.i.d assumption to help derive our formulation of GPD. Critically, in the real-world dataset, such as sonar heart, vehicle, fourclass, poker, codrna, usps, MNIST and CIFAR10, our method still works very well, and outperforms the competitors as validated in the experiments. The i.i.d assumptions of the observations are not necessarily always established in these datasets.This not only demonstrates the efficacy of our algorithm, but also empirically validates that our algorithm can be generalized to these general datasets. On the other hand, it is also quite common in statistics and machine learning that one algorithm is derived by some strong assumptions, but it works very well in practice. For example, the Naïve Bayes classifier is formulated by assuming very strong (naive) independence assumptions of features, while it works very well to general datasets. 

\section{Conclusion}
This paper introduces GPD $k$-means to improve $k$-means clustering ability by EVT, with a na\"ive baseline GEV $k$-means. 
We propose the concept of centroid margin distance, and use GPD to establish a probability model for each cluster, and perform clustering based on covering probability function derived from GPD. Extensive experiments on synthetic datasets and real datasets show that our GPD $k$-means outperforms competitors by clustering from the probabilistic perspective.

\section{Acknowledgement}
Sixiao Zheng, and Ke Fan are the co-first authors; and Yanxi Hou is the corresponding author. This work was supported in part by Nation Science Foundation of China Grant 71991471, National Science Foundation of Shanghai Grant 20ZR1403900, and the Science and Technology Commission of Shanghai Municipality Project (19511120700).






\bibliographystyle{IEEEtran}
\bibliography{GPD_kmeans}

\begin{IEEEbiography}[{\includegraphics[width=1in,height=1.25in,clip,keepaspectratio]{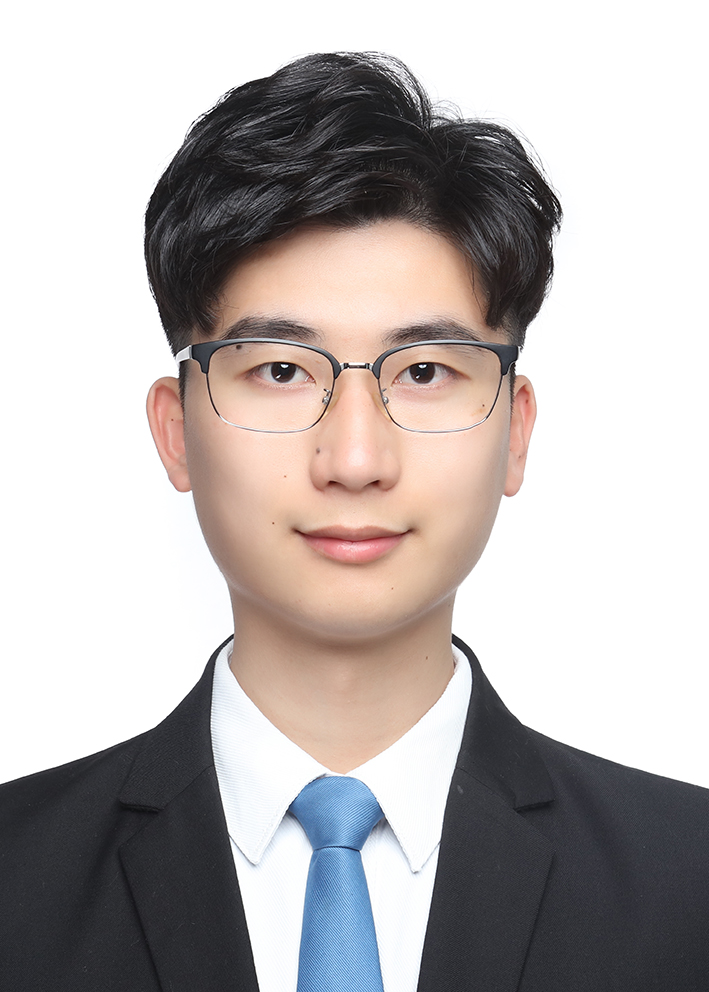}}]{Sixiao Zheng} received the B.E. degree in communication engineering from South China Normal University in 2018 and the M.S. degree in computer science from Fudan University in 2021. Since 2021, he has been a Researcher of Tencent, China. 
His research is focused on machine learning, incremental learning, semantic segmentation and object detection.
\end{IEEEbiography}



\begin{IEEEbiography}[{\includegraphics[width=1in,height=1.25in,clip,keepaspectratio]{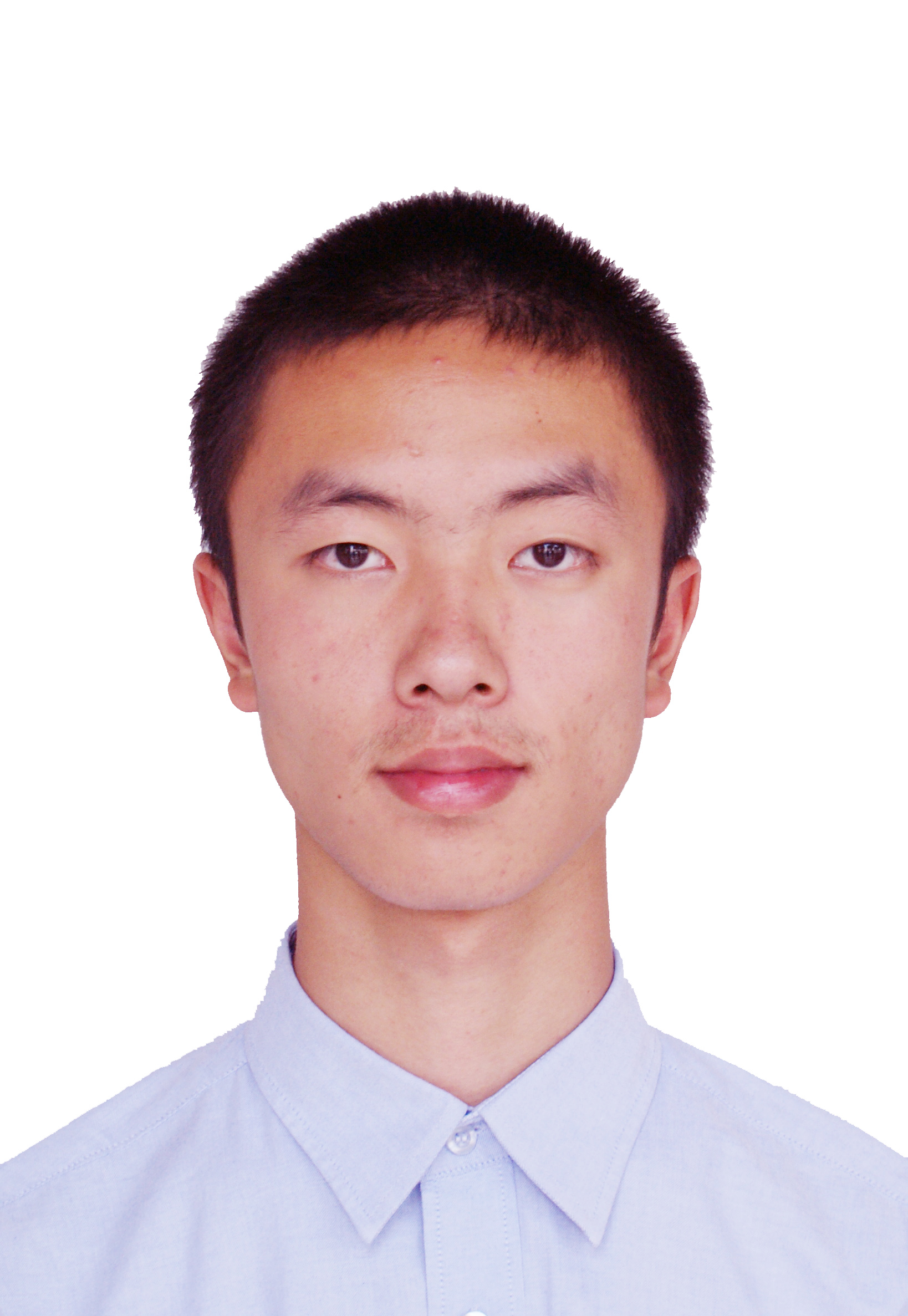}}]{Ke Fan} received the bachelor's degree from the School of Data Science, Fudan University. He works under the supervision of Professor Yanwei Fu. His current research interests are few-shot learning and unsupervised learning.
\end{IEEEbiography}

\begin{IEEEbiography}[{\includegraphics[width=1in,height=1.25in,clip,keepaspectratio]{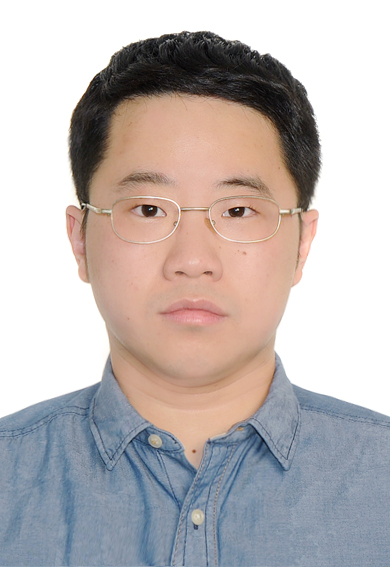}}]{Yanxi Hou} received both the B.S. and M.S. degrees in statistics from Fudan University, Shanghai, China, in 2013, the Ph.D. degree in mathematics from Georgia Institute of Technology, Atlanta, USA, in 2017. His research focuses on statistics and extreme value theory.
\end{IEEEbiography}

\begin{IEEEbiography}[{\includegraphics[width=1in,height=1.25in,clip,keepaspectratio]{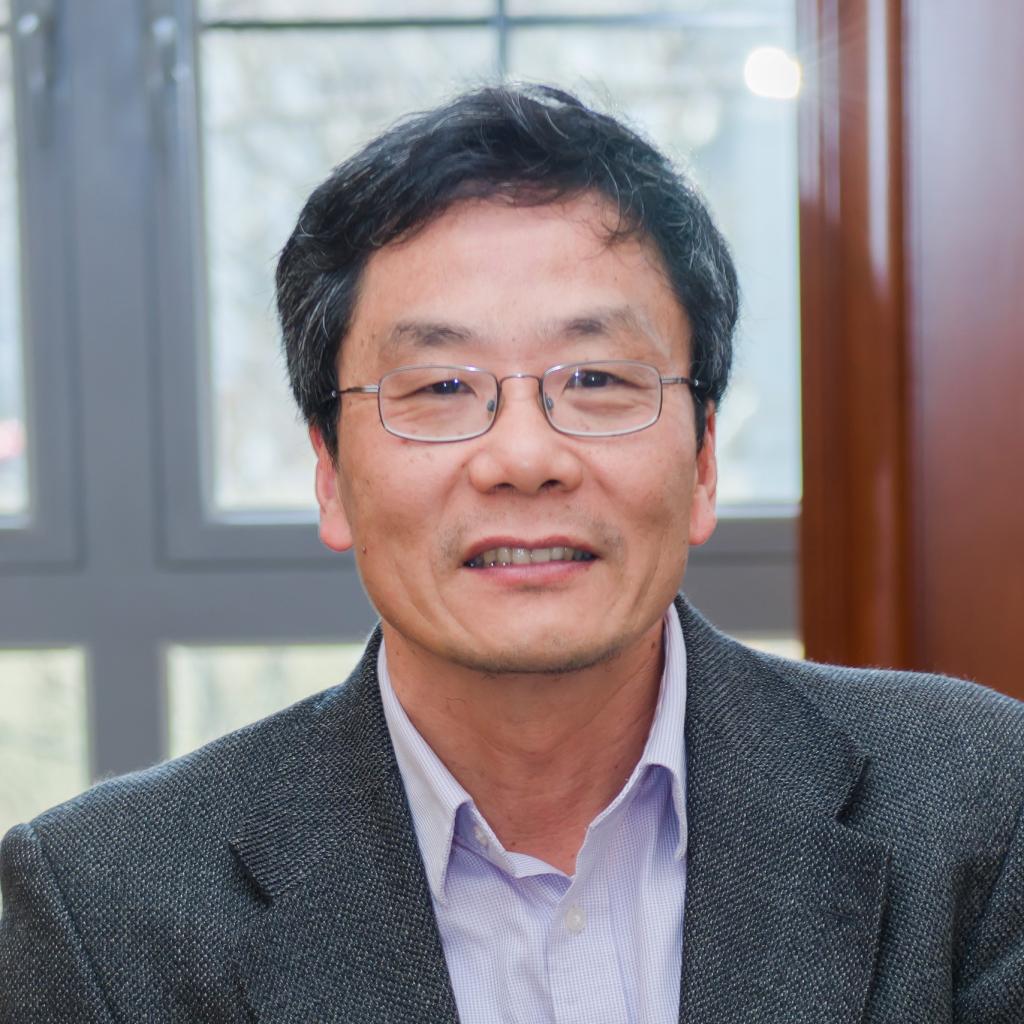}}]{Jianfeng Feng} received all his academic degrees from Peking University in mathematics, Peking, China, in 1993. He is the chair professor of Shanghai National Centre for Mathematic Sciences, and the Dean of Brain-inspired AI Institute and the head of Data Science School in Fudan University since 2008. He has been developing new mathematical, statistical and computational theories and methods to meet the challenges raised in neuroscience, mental health and brain-inspired AI researches. He was awarded the Royal Society Wolfson Research Merit Award in 2011, as a scientist ‘being of great achievements or potentials’. He was invited to deliver 2019 Paykel Lecture at the Cambridge University.
 \end{IEEEbiography}
 
\begin{IEEEbiography}[{\includegraphics[width=1in,height=1.25in,clip,keepaspectratio]{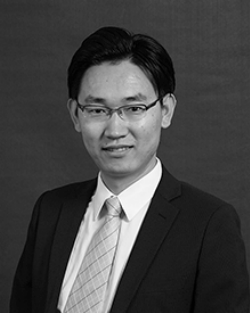}}]{Yanwei Fu} received the MEng degree from the Department of Computer Science and Technology, Nanjing University, China, in 2011, and the PhD degree from the Queen Mary University of London, in 2014. He held a post-doctoral position at Disney Research, Pittsburgh, PA, from 2015 to 2016. He is currently a tenure-track professor with Fudan University.  
His work has led to many awards, including the IEEE ICME 2019 best paper.
He published more than 80 journal/conference papers including IEEE TPAMI, TMM, ECCV, and CVPR. His research interests are one-shot learning, and learning based 3D reconstruction.
\end{IEEEbiography}

\end{document}